\documentclass[final,3p,times,twocolumn]{elsarticle}

\usepackage{hyperref}

\usepackage{amssymb}

\usepackage{graphicx}
\usepackage[cmex10]{amsmath}
\usepackage{amssymb}
\usepackage{algorithmic}
\usepackage{array}
\usepackage[table]{xcolor}
\usepackage{hhline}
\usepackage[caption=false,font=footnotesize]{subfig}
\usepackage{fixltx2e}

\usepackage{url}
\biboptions{numbers,sort&compress}
\usepackage{algorithm}
\usepackage{multirow}
\usepackage{bigdelim}
\usepackage{rotating}
\usepackage{color,ulem}
\usepackage{cases}
\usepackage{longtable}
\usepackage{pdflscape}

\newcommand{\blue}{\textcolor[rgb]{0.0,0.0,0.0}}
\def\eg{e.g. }
\def\etal{et al. }
\def\ie{i.e. }

\journal{}
\makeatletter
\def\ps@pprintTitle{%
 \let\@oddhead\@empty
 \let\@evenhead\@empty
 \def\@oddfoot{}%
 \let\@evenfoot\@oddfoot}
\makeatother

\begin{document}
\begin{frontmatter}
\title{Multiple Human Tracking in RGB-D Data: A Survey}

\author{Massimo Camplani, Adeline Paiement, Majid Mirmehdi, Dima Damen, Sion Hannuna, Tilo Burghardt, Lili Tao}
\address{Visual Information Laboratory \\ Faculty of Engineering \\ University of Bristol \\ Bristol BS8 1UB \\
~\\ Corresponding Author: Majid Mirmehdi \\ email: m.mirmehdi@bristol.ac.uk}

\begin{abstract}
Multiple human tracking (MHT) is a fundamental task in many computer vision applications. Appearance-based approaches, primarily formulated on RGB data,  are constrained and affected by problems arising from occlusions and/or illumination variations. In recent years, the arrival of cheap RGB-Depth (RGB-D) devices has {led} to many new approaches to MHT, and many of these integrate color and depth cues to improve each and every stage of the process. In this survey, we present the common processing pipeline of these methods and review their methodology based (a) on how they implement this pipeline and (b) on what role depth plays within each stage of it. We identify and introduce existing, publicly available, benchmark datasets and software resources that fuse color and depth data for MHT. Finally, we present a brief comparative evaluation of the performance of those works that have applied their methods to these datasets. 
\end{abstract}


\end{frontmatter}


\section{Introduction}\label{sec:introduction}

{Human tracking is a key component} in many computer vision applications, including video surveillance \cite{Wang20133}, smart environments \cite{multicameraMuseum}, assisted living \cite{cardinaux_video_2011}, advanced driver assistance systems (ADAS) \cite{geronimoADASreview}, and sport analysis \cite{playerTracking}. They are usually centered around RGB sensors and are characterized by a variety of limitations, such as occlusions due to cluttered or crowded scenes and varying illumination conditions. The vast literature landscape in this research area has widened even further in the last few years, due to the introduction and popularity of low-cost {RGB-Depth (RGB-D)} cameras (such as the Kinect \cite{xboxKinect} and Asus Xtion \cite{AsusXtion}). This has enabled the development of new algorithms that integrate depth and color cues to improve detection and tracking systems \cite{shottonReview}.

{The aim of this survey paper is to summarise and focus on the area of multiple human \textit{tracking from the combination of color (RGB) and depth (D)} data, given that \blue{cheap depth-enabled} sensors are becoming  ubiquitous in vision research labs. \blue{The survey is not however limited to methods using active sensing RGB-D devices, but also encompasses state-of-the-art passive sensing stereo-based human tracking techniques, where color and depth are again jointly relied upon to enable tracking.}

We do not review methods based only on RGB features as that would need a dedicated survey of its own and would demand much greater space - for RGB only MHT, the reader is referred to the reviews presented by Dollar \etal \cite{dollarReview} on color-based pedestrian detection and  Luo \etal for color-based multi-object tracking \cite{LuoZK14Review}. The intention here rather is to address and summarize an area that is now of far-reaching interest to a huge community of researchers.} 

Four main computer vision topics were identified in \cite{shottonReview} that could benefit from depth information: human activity analysis and recognition \cite{ChenSurveyDepthImagery,datasetRGBDreview}, hand gesture analysis \cite{gestureDepthREVIEW}, 3D mapping \cite{3dmappingCremers} and object detection and tracking. 
For example, the effect of occlusions can be reduced by using the 3D information contained in depth data, 
or more reliable features can be extracted in scenes undergoing illumination variations since such variations have low impact on depth sensors. 
Moreover, depth can be used to extract a richer description of the scene allowing to simplify the tracking problem, for example by adding physical constraints on human appearance and size. On the other hand, certain depth sensor characteristics, such as limited capture range, or scene characteristics, such as excessive natural light and reflective surfaces, reduce the reliability of depth data in some operating conditions, e.g. in outdoor scenarios. Color and depth data can be significantly complementary, and hence their efficient combination and processing can dramatically reduce the effect of the problems that affect them individually. 
In this survey, we focus on the analysis of algorithms, and available datasets and software, that \textit{combine color and depth data} for multiple human tracking.  
Most previous survey papers on human tracking do not provide such coverage and are limited to one or other aspects of MHT. For example, in \cite{Wang20133}, an in-depth review of surveillance systems is provided, with particular focus on challenges in using large camera networks. In \cite{monocularPedestrian}, pedestrian detection methods using color-based approaches are surveyed while the pedestrian detection review presented in \cite{geronimoADASreview} is mainly focused on ADAS systems. {The survey presented in \cite{dollarReview} proposes an extensive evaluation of sixteen pedestrian detectors that are based on a sliding window strategy. In \cite{crowdReview}, the focus of the survey is on algorithms for high-level crowd scene understanding. A review of human detection algorithms in video surveillance applications is presented in \cite{eurasip2013} where the main sub-modules of the human detection task are identified (object detection and object classification) and the state of the art algorithms are appraised by describing the different strategies used in each sub-module. The survey presented in \cite{Zhou20081} summarizes the advances in human body parts tracking for rehabilitation purposes.} {Table \ref{tab:survey} lists the recent surveys that are related in some fashion to MHT.} 

To the best of our knowledge, the surveys most closely associated to ours here are those presented in \cite{shottonReview,LuoZK14Review,ChenSurveyDepthImagery}. These cover similar themes but come with certain limitations. The work in \cite{shottonReview} reviews recent Kinect-based applications in computer vision, including a very brief survey of RGB-D based trackers. 
The review in \cite{ChenSurveyDepthImagery} is focused on the recent advances on human activity analysis using depth imagery, while the problem of human detection and tracking is only marginally discussed. 
Finally, in \cite{LuoZK14Review}, a general review of multiple object tracking is presented, but the analysis, dedicated to approaches that combine color and depth data, is {limited and brief}.

\begin{table}
\caption{{Recent} related surveys (most recent first)}
\label{tab:survey}
\vspace{0.2 cm}
\centering
\resizebox{\columnwidth}{!}{
\begin{tabular}{c|c|c}
{\bf Year} & {\bf Article} & {\bf Topic} \tabularnewline
\hline
\blue{2016} & \blue{Zhang \etal \cite{datasetRGBDreview}} & \blue{RGB-D dataset for action recognition}\tabularnewline
\hline
2014 & Luo \etal \cite{LuoZK14Review}  & Multiple object tracking\tabularnewline
\hline 
2013 & Chen \etal \cite{ChenSurveyDepthImagery}  & Human activity analysis \tabularnewline
\hline 
2013 & Han \etal \cite{shottonReview}  & Recent Kinect applications\tabularnewline
\hline 
2013 & Li \etal \cite{crowdReview}  & Crowd monitoring\tabularnewline
\hline 
2013 & Manoranjan \etal \cite{eurasip2013}  & Human detection in surveillance\tabularnewline
\hline 
2013 & Wang \cite{Wang20133}  & Multi-camera video surveillance\tabularnewline
\hline 
2012 & Dollar \etal \cite{dollarReview}  & Pedestrian detection \tabularnewline
\hline 
2010 & Geronimo \etal \cite{geronimoADASreview}  & ADAS systems\tabularnewline
\hline 
2009 & Enzweiler \etal \cite{monocularPedestrian}  & Pedestrian detection\tabularnewline
\hline 
2008 & Zhou \etal \cite{Zhou20081}  & Human tracking in rehabilitation\tabularnewline
\hline 
\end{tabular}
}
\end{table}

\normalsize
 
It is worth noting that we do not consider general single object trackers based on combined depth and color features, such as the recent works presented in \cite{garca_adaptive_2012,song_tracking_2013,wang_multi-cue_2014,Zhong2015710}, since they are more focused on the optimization of {appearance} and motion models rather than facing the specific challenges of MHT, \blue{or are concerned with tracking inanimate objects}. Furthermore, we do not include detection only methods, \eg \cite{Walk2010,Spinello2011,Wang2014}, and methods that use depth only for MHT, \eg \cite{Xia2011,Stahlschmidt2014,EPFL-CONF-206218,Fosty2013},
 or depth and reflectance, such as \cite{Dondi2013}. 

In summary, we provide here a review of the state-of-the-art on MHT algorithms that integrate depth and color data, characterizing them based on (a) trajectory representation and matching and (b) how they exploit depth information to improve various stages of the processing pipeline. We {also provide a review of the constraints of use of these algorithms}, {and we} examine existing online resources, i.e. benchmark datasets and source codes, and present a comparison of the very few such resources made available to the community. \blue{The audience of this survey is not limited to researchers working directly in the development of tracking algorithms, but also includes those who wish to employ a tracking method that is relevant to their application area, where colour and depth sequences are to be analysed, such as 
the very active research area of action recognition \cite{ChenSurveyDepthImagery,datasetRGBDreview}, smart environments \cite{smartMuseum}, health-care applications \cite{fallDetection,sphereMAGAZINE}, and applications mentioned in \cite{shottonReview}.}

Next, in Section \ref{sec:characterization}, we present the common processing stages of a typical MHT system, along with a variation on it employed by some works. Amongst other topics, we cover some generic descriptions of a person and introduce two characterizations of MHT systems based on their matching strategy and use of depth. These characterisations are then used in Sections \ref{sec:pipelines} and \ref{sec:depth_use} to 
survey state-of-the-art approaches. 
Practical issues, such as type of sensor, camera position, and speed of computation, are considered in Section \ref{sec:applications}. Section \ref{sec:resources} presents an overview of online datasets and software resources for RGB-D MHT. We then compare existing evaluations derived from some of the works in this survey in Section \ref{sec:evaluation} and conclude  in Section \ref{sec:conclusion}.

\section{Multiple People Detection and Tracking Techniques in RGB-D Data}
\label{sec:characterization}

In this section, we identify the main approaches to MHT from combined color and depth data. We first present the processing pipeline that can be attributed to the greater set of works in the literature 
{and then characterize the works we review based on}
(a) which trajectory representation is used and its matching,
and (b) how and for which purpose depth data is exploited. 

In MHT, detections of multiple people {are} normally aggregated into independent tracks, one for each person, in order to establish
their respective trajectories. Tracks may contain position, motion, and appearance descriptions. We shall use the words `track' and `trajectory' interchangeably in the rest of this article.

The common processing pipeline is illustrated in Fig.~\ref{fig:pipelines}.
MHT methods normally perform the stages indicated by the solid lines in Fig.~\ref{fig:pipelines}, with first a detection stage that searches for occurrences of humans in a new frame, based on a generic description of a {person} (elaborated later below). It may possibly be preceded by an optional Regions of Interest (ROIs) selection stage (the dashed-line box in Fig.~\ref{fig:pipelines}), that allows for the reduction of the search space. Then, a matching step associates these new detections to {the} trajectories {based on a matching strategy and a similarity measure, computed from position and, more often than not, appearance.

\begin{figure}
\centering
\includegraphics[width=1.0\linewidth]{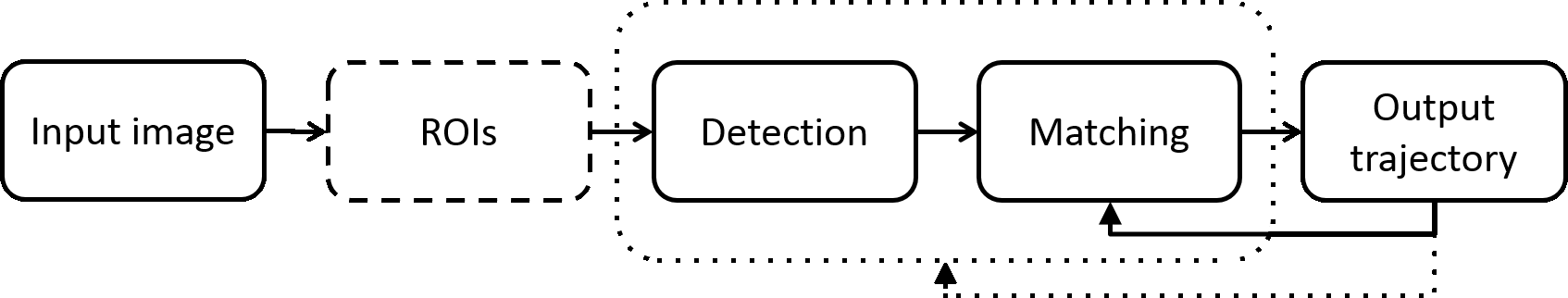}
\caption{Common processing pipeline for MHT - the dashed-line stage is an optional step of the pipeline, while the dotted rectangle and arrow depict a variation of it.}
\label{fig:pipelines}
\end{figure}
There are numerous approaches to performing the matching process.
These  rely on the active trajectories to provide (i.e. effectively feedback) a representation of the target's motion and appearance to their matching stage (the solid arrow in Fig.~\ref{fig:pipelines}).
The pool of active trajectories is managed by the matching stage, with new trajectories created when detections cannot be associated to the existing ones, and old trajectories discontinued when certain termination criteria are {met.}

In a variation to the common pipeline, depicted by the dotted line and box in Fig.~\ref{fig:pipelines}, the detection and matching stages may be directed by trajectories and their representations rather than by a generic representation of a person. Thus, currently tracked people are directly detected at the {position} predicted by their trajectory representation's motion model in a significantly reduced search space.
In effect, this amounts to combined detection and matching.
This variation of the MHT processing pipeline still requires a generic person description for initializing new trajectories by detecting people that are not yet tracked. 
Note that some methods also use a generic person description in the combined detection and matching stage in addition to the trajectory representation, in order to ensure that tracks do not switch to background objects of similar {predicted position} and appearance to that of the target.

{Section \ref{sec:pipelines} provides a detailed description of implementations of the MHT pipeline (and its variation), including comparing different fulfilments of the matching stage.}
It should be stressed that both the main pipeline and its variation are by no means specific to RGB-D based methods, and the same can apply to MHT methods based on {RGB data} only.

Trajectory representations in MHT methods vary significantly between implementations as well, as illustrated in Fig.~\ref{fig:categories_representation}.
{Thus, {although} all reviewed methods employ a motion model in their trajectory representation, the use of an appearance model (\eg color histogram, texture, joint {RGB} and depth feature, etc...) is optional, as represented by the dashed arrow (or blue sub-tree) in Fig.~\ref{fig:categories_representation}.}
Both motion and appearance models may be built from an observation in a single frame, or from richer information that accounts for the history of the target. 

\begin{figure}
\centering
\includegraphics[width=0.9\linewidth]{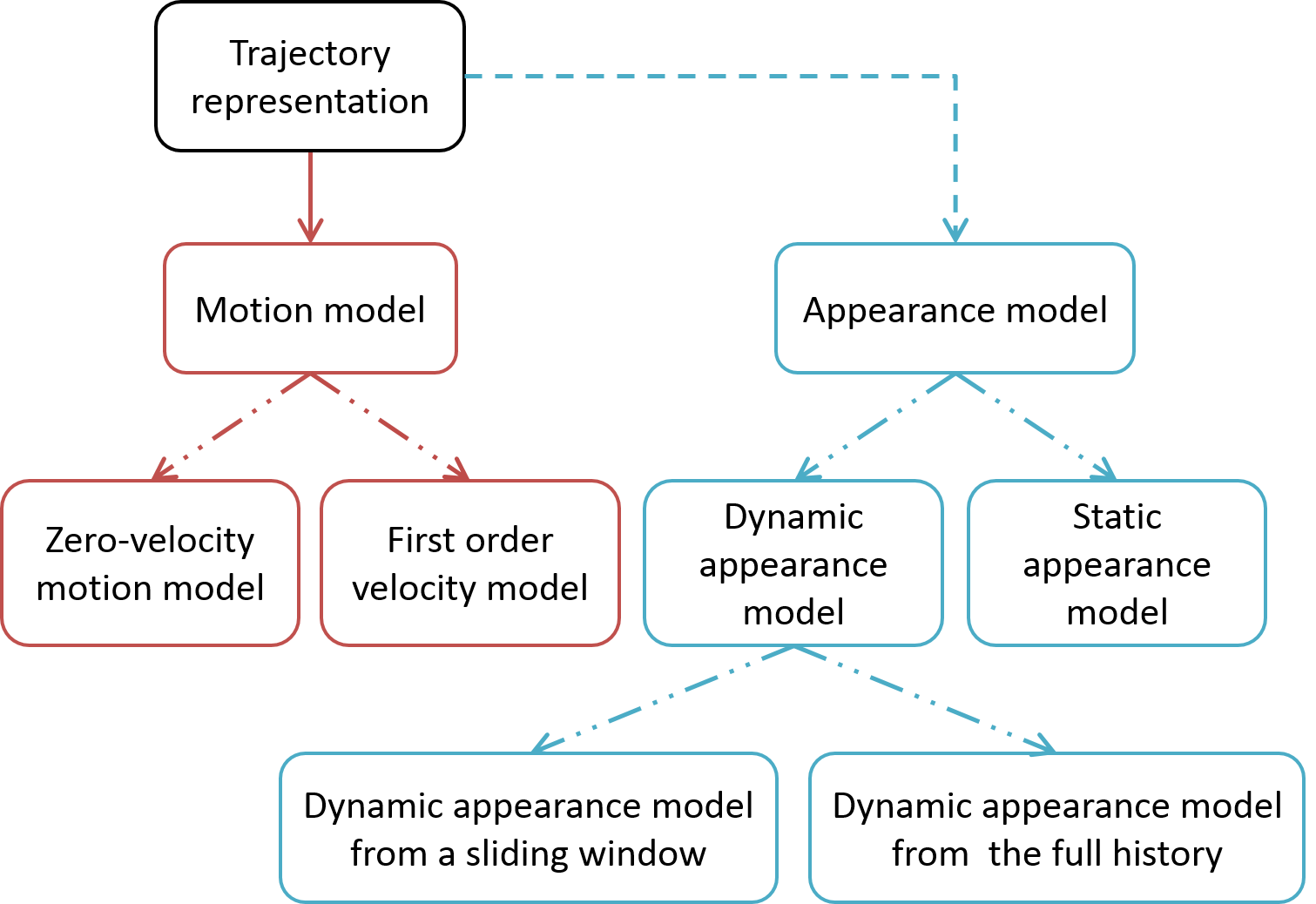}
\caption{Categorisation of the different models that make up the trajectory representation used for matching. The dashed arrow denotes an optional model for the trajectory representation, while semi-dotted arrows indicate where one or the other of two possibilities is selected.}
\label{fig:categories_representation}
\end{figure}

Two types of motion models may be identified. The first, denoted as 'zero-velocity motion model', assumes stationary position of the target, while the second describes their velocity, yielding a first order characterization of their movements. Higher order motion models, such as one that includes the target's acceleration, would be equally possible, but are not addressed in this survey as no methods in RGB-D MHT were encountered that employed them.
Static appearance models may be built from one or a few initial frames 
and remain fixed for the duration of the trajectory's lifetime, while dynamic appearance models may be derived from all previous observations of the target or from a sliding window. Such models are updated as new observations become available, in order to account for varying appearances, {due to different body orientation relative to the sensor or changing illumination conditions}. Yet these dynamic models could result in incorrect descriptions in case of failure in tracking, such as drifting.
MHT methods may use any combination of these different possible (static and dynamic) motion and appearance models.

Depth data can be exploited to enhance {RGB-based} MHT. The methods that we review can be characterized by how and {at} which stage {of} the MHT pipeline they employ depth information. Indeed,
depth may support each and every stage of the MHT pipeline, as indicated in Fig.~\ref{fig:categories_depth}. It can help to specify \textit{ROIs} in the image corresponding to 3D physical scene regions of significance, e.g. a doorway or {passage}, to help reduce the search space for the detection stage (left branch of Fig.~\ref{fig:categories_depth}). 
{Depth information} may also {increase the robustness of} \textit{human detection}, {by enhancing the generic description of a person with 3D shape information
(middle branch of Fig.~\ref{fig:categories_depth})}.
Finally, depth can help in \textit{matching} detected candidates and trajectories (right branch of Fig.~\ref{fig:categories_depth}),
by {providing the information needed to track people in 3D, and by further
enriching the appearance descriptions of people, that are traditionally based on {RGB} information only.} {The various uses of depth information in published work} will be detailed in Section \ref{sec:depth_use}.

\begin{figure}
\centering
\includegraphics[width=0.9\linewidth]{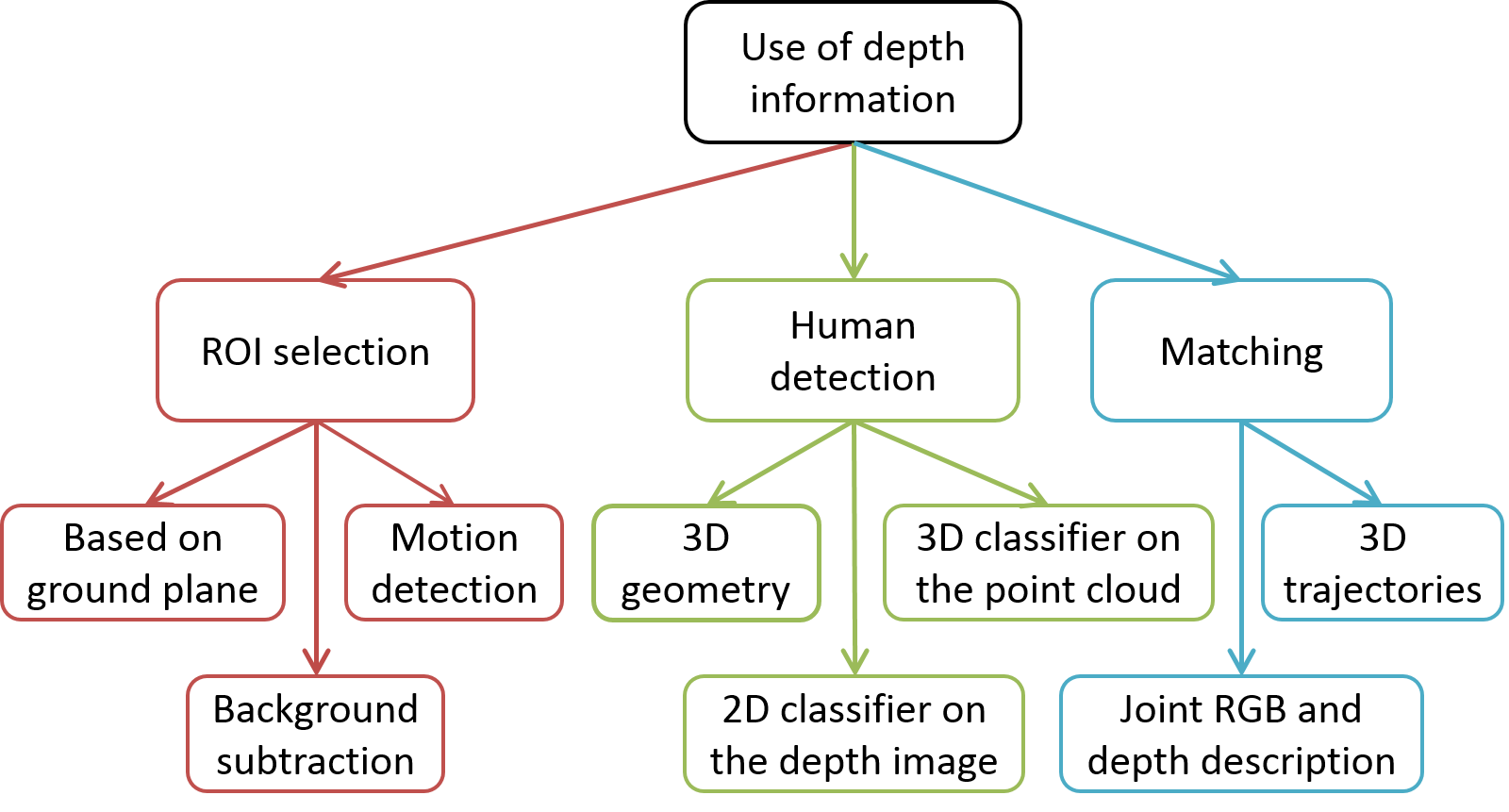}
\caption{Categorisation of the uses of depth information in MHT methods from RGB-D data.}
\label{fig:categories_depth}
\end{figure}

The \textit{generic description of a person} that drives the detection stage, is often made up of a number of RGB and depth cues. Then, in the detection stage, a cascade of RGB and depth based descriptors is applied to either the full image or ROIs, starting with the less computationally expensive ones, which are generally depth based descriptors of the human shape. 
When using RGB information, the generic representations for a person often takes the form of an Histogram of Oriented Gradients (HOG) \cite{peopleHOGdalalTriggs} descriptor of the full or upper-body.
Other examples of possible generic person descriptions from RGB data are provided by the poselet-based human detector of \cite{Bourdev2009}, {the \blue{deformable} part-based models of \cite{lsvm-pami} \blue{(DPM)}} or use the Viola and Jones Adaboost cascade \cite{violaJones}.
Table \ref{tab:generic_descriptions_person} summarizes the different generic descriptions of people used by the various methods reviewed here.}

\begin{table}[!ht]
\caption{Types of generic descriptions of a {person}.} 
\label{tab:generic_descriptions_person}
\vspace{2mm}
\centering
\resizebox{0.94\columnwidth}{!}{
\begin{tabular}{>{\centering}m{0.38\linewidth}|>{\centering}m{0.165\linewidth}|>{\centering}m{0.48\linewidth}}
{\bf Method} & {\bf Depth descriptor} & {\bf RGB  descriptors} \tabularnewline \hline \hline
Bansal \etal \cite{Bansal2010} & \checkmark & 2D contour matching + HOG \tabularnewline \hline
Salas \& Tomasi \cite{salas_people_2011} & -  & HOG \tabularnewline \hline
Dan \etal \cite{Dan2012} & \checkmark & - \tabularnewline \hline
{Darrell \etal \cite{darrellIJCV}} & - & Face and skin detector \tabularnewline \hline
Han \etal \cite{han_employing_2012} & \checkmark & - \tabularnewline \hline
{Bajracharya \etal \cite{Bajracharya2009}} & {\checkmark} & - \tabularnewline \hline
Zhang \etal \cite{zhang_real-time_2013} & \checkmark & HOG + poselet \tabularnewline \hline
{Galanakis \etal \cite{Galanakis2014}} & {\checkmark} & - \tabularnewline \hline
Liu \etal \cite{icipColorHeightHist,Liu201516,LiuNEWDETECTOR} & \checkmark & Joint rgb \& height histogram \blue{+ physical priors \cite{LiuNEWDETECTOR}} \tabularnewline \hline
Luber \etal \cite{luberIROS11} and \\ {Linder \etal \cite{linderFUSION14}}& \checkmark & HOG \tabularnewline \hline
{Ess \etal \cite{Ess2009}} & {\checkmark} & {HOG based detectors} \tabularnewline \hline
Jafari \etal \cite{aachenTracker} & \checkmark & HOG \tabularnewline \hline
Mu\~{n}oz-Salinas \etal \cite{Munoz2007} & - & Face detector \tabularnewline \hline
Munaro \etal \cite{munaro_tracking_2012,Munaro2014} & \checkmark & HOG \tabularnewline \hline
Almaz\`{a}n and Jones \cite{Almazan2014,almazan_tracking_2013} & \checkmark & - \tabularnewline \hline
{Bahadori \etal \cite{bahadori2005}} & - & Temporal Color based model \tabularnewline \hline
{Beymer \etal \cite{beymer1999}} & \checkmark & - \tabularnewline \hline
{Satake \etal \cite{satake2013}}& \checkmark & SIFT \tabularnewline \hline
{Vo \etal \cite{Vo2014}} & \checkmark & HOG + face detector \tabularnewline \hline
{Harville \etal \cite{Harville2004127}} & \checkmark & - \tabularnewline \hline
Mu\~{n}oz-Salinas \etal \cite{MunSal2008,munozsalinaspeople2009} & \checkmark & - \tabularnewline \hline
Mu\~{n}oz-Salinas \etal \cite{Munoz2008b} & \checkmark & Adaboost classifier for upper body + ellipse fitting at head location \tabularnewline \hline
Choi \etal \cite{choi_detecting_2011,pamiSavarese} & \checkmark & HOG + face detector + motion detector + skin color recognition \tabularnewline\hline
{Migniot \etal \cite{Migniot2014}} & \checkmark & - \tabularnewline \hline
{Gao \etal \cite{gaoIEEEITS}} & - & HOG + \blue{DPM} \tabularnewline \hline
\blue{Ma \etal \cite{DPMMultiple}} & - & \blue{HOG + DPM} \tabularnewline \hline
\end{tabular} 
}
\end{table}

{Next, in Section \ref{sec:pipelines}, we review all  RGB-D MHT methods known to us, leveraged on how they implement the MHT pipeline. We  characterize these works based on {their applied trajectory representation and matching strategy,} following the categorization proposed in Fig.~\ref{fig:categories_representation}. Then, in Section
\ref{sec:depth_use}, we again review and characterize these \textit{same methods} based on their adoption of depth information, describing the uses of depth for each stage of the MHT pipeline,
according to Fig.~\ref{fig:categories_depth}.}

\section{Survey by MHT Pipeline Implementation}
\label{sec:pipelines}

This section details how the pipeline for MHT, described in Section \ref{sec:characterization}, 
has been implemented, including optional stages and variations.
The emphasis is on (complexity of) trajectory representation (see Fig.~\ref{fig:categories_representation}) and matching. We may refer to depth data in this section for some of the works - the details of their use of depth will be provided in Section \ref{sec:depth_use}.

\subsection{Implementations of the main pipeline}
\label{sub:sequential_det_track}

We encountered only {four} works that build their trajectory representation exclusively from the previous frame \cite{darrellIJCV,Bansal2010,Dan2012,salas_people_2011}. The principle characteristics of their implementation of the MHT pipeline are indicated in the first {four} rows of Table \ref{tab:implementation_pipeline}. 

{Darrell \etal \cite{darrellIJCV} present a stereo-based tracking approach using the target's position and size constancy from frame to frame. In particular, candidates are detected by using a segmentation approach that allows to identify connected component in the disparity images corresponding to regions in the 3D space with a typical volume occupied by a person facing the camera. For each detected region, a cascade of face and skin detectors, and geometric constraints, are applied to validate the target's head position. A long term model is generated by considering skin and face average color, appearance color histogram, face pattern and height extracted from depth data. These features are used to solve occlusions and target re-identification in case of targets re-entering in the scene.} 

Bansal \etal \cite{Bansal2010} first detect people 
after an ROI selection stage, 
{using a combination of depth cues, and a HOG detector that is applied to a selection of {edges obtained} by preliminary template matching with several 2D contours of different body parts.}
Then, {they match detections with trajectories from} the previous frame by image patch-correlation. This is performed in the area of the image that contained the previous observation of the person, after correction for camera motion estimated by visual odometry. Thus, the trajectory representation is made up of a zero-velocity motion model in the 2D image {coordinates} and  an appearance model that consists of an image patch around the detection in the previous frame. This amounts to a dynamic appearance model built from a sliding-window of one frame-width.

Salas and Tomasi \cite{salas_people_2011} detect and track all objects in ROIs that denote foreground, and then they select the paths that have, at some point in time, a detection with a high confidence score from a HOG based human detector. The {matching stage} is performed by dynamically building a directional connected graph
of the foreground object detections. These are organised into layers that correspond
to the frames they originate from, and they are interconnected by the graph's edges
in chronological order. The cost of an edge is the probability for its two nodes
(or foreground detections) {belonging} to the same object. Based on these costs,
the tracks are selected as the most strongly connected paths in the graph.
A greedy algorithm is used for extracting individual paths, starting from the oldest remaining
detection, and selecting the strongest connection locally between two adjacent node layers. 
The edge cost, used for matching, is estimated
from the similarity of the color signatures, measured using the Earth Mover's Distance \cite{Rubner2000}, and from the distance between the 3D locations of both detections that is expected to be proportional to the elapsed time.
Thus, the trajectory representation consists of a zero-velocity motion model, and an appearance model made up of the color signature \cite{Rubner2000} of the blob in the previous frame.

Dan et al. \cite{Dan2012} use depth {information} for detection.
{All detected candidates are then} matched independently to detections in the previous frame by maximizing a score that assesses both appearance {similarity and} closeness in 3D space. 
The trajectory representation used for matching is made up of a
RGB-D based dynamic appearance model with a sliding window of one frame, 
along with a zero-velocity motion model.
A backward/forward matching strategy {is used}, where all detections in  frame $t$ are matched to those in frame $t-1$ (backward matching), and vice versa (forward matching), which allows handling trajectory splits and merges, which
may arise from the failure of detection in one {direction} that may match two people against the same candidate. 

All {four} methods in \cite{darrellIJCV,Bansal2010,Dan2012,salas_people_2011} propose a crude motion model that does not describe a person's movements sufficiently well,  although \cite{salas_people_2011} expects the distance travelled to be proportional to time. The movement itself, and in particular its direction, are not captured by the trajectory representation. {Thus, these methods} 
are more likely to suffer from incorrect identifications when a track `jumps' from one person to another, and from wrong detections being integrated into the tracks.
In addition, both their motion and appearance models are made from the observations in the previous frame only. Hence, in case of occlusion, a person cannot be tracked any longer and the associated trajectory is automatically discontinued. A new, independent trajectory would have to be created if the person {re-emerges}.

The methods we present in the rest of this, and the next subsection, occupy rows {5} to the end of Table \ref{tab:implementation_pipeline}. These address the above issues (a) by proposing motion models that describe the motion of the target to the first order, and (b) by {building appearance models from richer temporal information, which allow} for maintaining consistent trajectory representations, and help prevent the model from changing dramatically in cases of temporary detection failure over a few frames. 

\begin{table*}[!ht]
\caption{Characterization of the methods based on their MHT pipeline implementation. The number of frames indicated in the column 'Dynamic -- sliding window' indicates the width of the window. For Liu et al. [38], 
it is not known if the appearance model is static or dynamic.}
\label{tab:implementation_pipeline}
\centering
\vspace{0.2 cm}
\resizebox{1\textwidth}{!}{
\begin{tabular}{c|>{\centering}p{1.3cm}|>{\centering}p{1.7cm}|>{\centering}p{1.3cm}|>{\centering}p{1.6cm}|c|>{\centering}p{2.0cm}|>{\centering}p{1.6cm}}
\multirow{2}{*}{\bf Method} & \multirow{2}{1.4cm}{\centering{} \bf ROI selection} & \multirow{2}{1.9cm}{\centering{}\bf Pipeline variation} & \multicolumn{2}{c|}{\bf Motion model} & \multicolumn{3}{c}{\bf Appearance model}\tabularnewline
\cline{4-8} 
 &  &  & {\bf Zero velocity} & {\bf 1st order velocity} & {\bf Static} & {\bf Dynamic - sliding window} & {\bf Dynamic - full history}\tabularnewline
 \hline \hline
Bansal \etal \cite{Bansal2010} & \checkmark & & \checkmark & & & \checkmark (1 frame) & \\ \hline
Salas \& Tomasi \cite{salas_people_2011} & \checkmark & & \checkmark & & & \checkmark (1 frame) & \tabularnewline 
\hline 
Dan \etal \cite{Dan2012} && & \checkmark & & & \checkmark (1 frame) & \tabularnewline 
\hline 
{Darrell \etal \cite{darrellIJCV}}  & {\checkmark} & & {\checkmark} &  & & \checkmark & \checkmark \tabularnewline 
\hline \hline 
{Bajracharya \etal \cite{Bajracharya2009}} & {\checkmark} & & & {\checkmark} & & {\checkmark (1 frame)} & \tabularnewline 
\hline 
Zhang \etal \cite{zhang_real-time_2013} & \checkmark & & & \checkmark & & & \checkmark  \tabularnewline 
\hline 
{Galanakis \etal \cite{Galanakis2014}} & {\checkmark} & & & {\checkmark} & & & {\checkmark}  \tabularnewline 
\hline 
Liu \etal \cite{icipColorHeightHist,Liu201516} & \checkmark & & & \checkmark & ? & ? & ?  \tabularnewline 
\hline 
Luber \etal \cite{luberIROS11} and  &  & & &  &  &  & \tabularnewline  
{Linder \etal \cite{linderFUSION14}}& & & & \checkmark & & & \checkmark \tabularnewline 
\hline 
{Ess \etal \cite{Ess2009}} & {\checkmark} & & & {\checkmark} & & & {\checkmark}  \tabularnewline 
\hline 
Jafari \etal \cite{aachenTracker} & \checkmark & & & \checkmark & & & \checkmark \tabularnewline 
\hline 
Mu\~{n}oz-Salinas \etal \cite{Munoz2007} & \checkmark & & & \checkmark & & & \checkmark  \tabularnewline 
\hline 
Munaro \etal \cite{munaro_tracking_2012,Munaro2014} & \checkmark & & & \checkmark & & & \checkmark \tabularnewline 
\hline
{Almaz\`{a}n and Jones \cite{Almazan2014}} & {\checkmark} & & & {\checkmark} & & {\checkmark (1 frame)} &\tabularnewline 
\hline 
{Bahadori \etal \cite{bahadori2005}}  & {\checkmark} & & & {\checkmark} & &  & \checkmark \tabularnewline 
\hline 
{Beymer \etal \cite{beymer1999}}  & {\checkmark} & & & {\checkmark} & &  & \checkmark \tabularnewline 
\hline 
{Satake \etal \cite{satake2013}}  & {\checkmark} & & & {\checkmark} & & {\checkmark} (30 frames) & \tabularnewline 
\hline 
 {Vo \etal \cite{Vo2014}}  & {\checkmark} & & & {\checkmark} & &  &   \tabularnewline 
\hline 
 {Harville \etal \cite{Harville2004127}}  & {\checkmark} & & & {\checkmark} & &  &  \checkmark \tabularnewline 
\hline  \hline
Almaz\`{a}n and Jones \cite{almazan_tracking_2013} & \checkmark & \checkmark & & \checkmark & & & \checkmark  \tabularnewline 
\hline 
Mu\~{n}oz-Salinas \etal \cite{MunSal2008} & \checkmark & \checkmark & & \checkmark & & & \checkmark \tabularnewline 
\hline 
Mu\~{n}oz-Salinas \etal \cite{munozsalinaspeople2009} & \checkmark & \checkmark & & \checkmark & & & \tabularnewline 
\hline 
Mu\~{n}oz-Salinas \etal \cite{Munoz2008b} & & \checkmark & & \checkmark & & & \checkmark \tabularnewline 
\hline 
Choi \etal \cite{choi_detecting_2011} & \checkmark & \checkmark & & \checkmark & & &  \tabularnewline 
\hline 
Choi \etal \cite{pamiSavarese} & \checkmark & \checkmark & & \checkmark & \checkmark & & \tabularnewline 
\hline 
{Migniot \etal \cite{Migniot2014}}  & {\checkmark} & & & {\checkmark} & &  &   \tabularnewline 
\hline 
{Gao \etal \cite{gaoIEEEITS}}  & {\checkmark} & & & {\checkmark} & &  &  \checkmark  \tabularnewline \hline

\blue{Ma \etal \cite{DPMMultiple}}  & {\checkmark} & & & {\checkmark} & & \checkmark &   \tabularnewline \hline
\end{tabular}
}
\end{table*}

In the work of Han \etal \cite{han_employing_2012}, the motion model determines target's velocity approximately by the mean and variance of its depth variations in the past 10 frames.
Their static appearance model is made up of color and texture histograms for the torso and legs, generated at the first observation of a new person,
with the torso and leg locations being detected using depth information. 
This trajectory representation is kept after the person leaves the scene, in order to allow re-identification in case of re-entry.
People are first detected in ROIs, as objects within a pre-defined height range appearing {for} a number of successive frames, based on depth information. Their best matching trajectory is selected from a linear combination of the appearance similarity and the continuity of
the depth variation. The former is assessed with the Bhattacharyya distance measure and the latter is expected to follow a Gaussian distribution {with a  mean and variance provided by the motion model,} under the assumption of {a constant speed}.

In \cite{Bajracharya2009}, Bajracharya \etal assume a target velocity of 2ms${}^{-1}$ in any one direction, hence the motion model does not depend on the data. 
The appearance model of the trajectory representation is made up of the color histogram of the last observation for the track. Matching is performed by comparing candidates, detected from depth information in ROIs, to trajectories, based on the color histograms of {the candidate and of the appearance model of the track}, evaluated by the Bhattacharyya measure. {Only trajectories that are predicted to be located close to the candidate's are considered.} 

In all other {RGB-D} MHT methods reviewed next which apply the main MHT pipeline, motion is modelled as the position and velocity of the tracked person from the previous frame. The position, and sometimes the velocity, of the next observation {are predicted from the model, and} then compared with the positions of new detections during the matching stage. {With the exception of \cite{Galanakis2014,Ess2009}, the methods reviewed next carry out their predictions using Kalman filtering.}

Some works find the best {association of a detected candidate to a track} independently for each detection or track. For example, in \cite{zhang_real-time_2013}, Zhang \etal find people in ROIs using a cascade of RGB and depth based detectors, where detected candidates from depth cues are verified by a HOG detector, and by the poselet-based human detector of \cite{Bourdev2009} that detects body parts. This last detector is rather computationally expensive, hence it is only applied to detected candidates that cannot be associated with existing targets in the matching stage.
The matching stage locates the best matching track {or static background object for each new detected candidate, using} a
Directed Acyclic Graph (DAG) {to handle the decision process.} 
{The DAG} performs {coarse matching} by position similarity first and then {finer matching} to account for appearance similarity. The appearance is
represented by a dynamic model, updated online by an AdaBoost algorithm. A classifier is trained by AdaBoost from weak nearest neighbor classifiers and color histogram features, with positive and negative examples taken from previous observations of the target and of other people and objects, respectively. This model is kept after the person leaves the scene to enable future re-identification.

Similarly in \cite{Galanakis2014}, Galanakis \etal model motion as the {target's} speed, computed between the last two frames, and use it to predict the next position of the target, assuming a constant velocity. Following the matching strategy of \cite{Argyros2004}, candidate detections, found by background subtraction, are associated with their nearest neighbor trajectories. Unlike \cite{Argyros2004} however, the distance to a trajectory combines the 3D distance to its predicted position and the appearance similarity, quantified as in \cite{Padeleris2013} by a correlation metric. The appearance model comprises the hue and saturation histograms of the upper and lower body which are found by reference to the depth data. It is updated by linear combination of the current model and the new histogram.
Liu \etal \cite{icipColorHeightHist,Liu201516} detect all candidate people in ROIs of a new frame from RGB-D {data} and then, {for each track, select the best detected candidate} by maximizing a correspondence likelihood that is a linear combination of distance
to the predicted position {and appearance} similarity, assessed by the Bhattacharyya measure. The appearance model of the trajectory representation is a joint color and height histogram. 
The authors do not give any indication whether their appearance model is updated. In order to handle {short-term} occlusions, the trajectory {is only terminated after 10 seconds of being lost}.

Other works consider all possible {associations of detections to tracks} in order to find a global optimisation that takes into account possible interactions between tracks, such as crossing of trajectories 
and sharing of detections. In \cite{luberIROS11}, Luber \etal 
build a tree of {association} hypotheses in a Multi-Hypothesis Tracker (MHT) framework, where matching probabilities, for all past and current frames, are computed from closeness to position and velocity predictions, and from appearance similarity. The MHT grows a hypothesis tree, pruned to the k-best hypotheses at each iteration in order to prevent exponential growth of the tree. The current best hypothesis, that jointly describes all tracks, is then selected at each frame, following \cite{Cox1996}. Similarly to \cite{zhang_real-time_2013}, the appearance model relies on a color and depth Adaboost classifier. {Linder \etal \cite{linderFUSION14} propose an extension of the method in \cite{luberIROS11} for group tracking. In particular, to characterize group movements, they add to the MHT framework a set of coherent motion indicators, such as relative spatial distance, difference in velocity, and difference in orientation of two given tracks.}

{Beymer	\etal \cite{beymer1999} propose a combination of stereo-based background subtraction (see \cite{EvelandStereoBG}) and a full body binary template to identify candidate targets. The binary template size is chosen according to the mean depth value of the foreground blob. A Kalman filter with a constant velocity model is used for tracking. A target's representation includes 3D space coordinates and two appearance models, a color model and the average disparity. These models are linearly updated taking into account the confidence rate of the person detector module, such that it introduces a smoothing factor in the update process, hence limiting the models' drift.} {A similar approach was proposed by Bahadori \etal \cite{bahadori2005},  using detected foreground regions and geometric constraints in their stereo setup to identify blobs containing candidate targets. For each blob, a fixed resolution and adaptive color based appearance model is generated, with each pixel  modeled as a unimodal distribution in the color space. Tracking is also performed with a Kalman filter, with a constant velocity model that takes into account the 3D depth position of the target and its appearance. The matching strategy is based on the minimization of the distance, considering both position and appearance, between all the detected candidates and the current active tracks. The generation of new tracks and the termination of lost ones is managed by a finite state machine system.} 

Ess \etal \cite{Ess2009} detect people in a Bayesian network  that accounts for the probabilities of human presence, as output by a color based detector, {given} the scene geometry and a generic person geometry description, both provided by depth data. Areas around the next expected target locations also see their detection likelihood increased. 
Then, they also {build multiple candidate tracks, from forward and
backward matching hypotheses, following \cite{aachenTrackerModule}'s  tracking framework. These hypotheses are generated from position predictions by a constant
velocity model and from appearance similarity measured using the Bhattacharyya distance on color histograms. The best tracks are selected, while enforcing that each person detection can 
only be matched to one trajectory.}
The trajectory's appearance model used for matching is {the mean} color histogram of all previous observations of the tracked person.
{Jafari \etal \cite{aachenTracker} use the same matching stage and trajectory representation. They perform detection} in ROIs based on depth at a close range and using a HOG detector \cite{aachenDetectorHOGGPU} in the far range.

{ Satake \etal \cite{satake2013} detect people by applying a classifier cascade  to the RGB-D data. First, a set of three binary templates \cite{satake2009robust}, containing frontal and side views of head and shoulders, are used to identify candidate regions in the disparity map. These are then validated and refined with an SVM classifier trained on HOG features to detect humans. An Extended Kalman filter is used to track the target in the 3D space. SIFT features \cite{SIFT} of the target are periodically collected to build an appearance model. Association between tracked targets and current frame detections is performed by thresholding on the number of matching SIFT features. }

Mu\~{n}oz-Salinas \etal \cite{Munoz2007} detect people from a face detector applied in ROIs selected from depth information. The face detector may suffer from false negatives in non-fronto-parallel views, therefore it is only applied at the very end of the detection cascade, and only to detected candidates that cannot be associated with existing targets in the matching stage. The matching stage finds the globally optimal associations of detected candidates to existing tracks using the Hungarian method \cite{Kuhn1955}. The matching likelihoods are computed from the distance to the predicted position and the similarity to {the color} histogram appearance model estimated with the Bhattacharyya measure. This model is updated by linearly combining its current values and the new observed color histogram. The track is discontinued if 
new observation of the target are not encountered after a time-limit.
Almaz\`{a}n and Jones \cite{Almazan2014} also use the Hungarian method to match candidates, detected from motion and size using depth information, 
to trajectories. The correspondence likelihood is based on the distance to the predicted position and on appearance similarity, evaluated using the Bhattacharyya measure. The appearance model combines a height histogram and the color distributions of its bins, and it is updated every 10 frames by replacing bins and their associated distributions by newly observed ones if available, i.e. if no occlusion happens. 

{Another method based on the Hungarian algorithm for matching  detected and tracked objects is that of Vo \etal in \cite{Vo2014}, where the authors identify background areas with a depth-based occupancy grid system. Candidate targets' search space is limited to the foreground areas which is analyzed with a cascade of classifiers, comprising face and skin detectors (see \cite{VoFaceDetector} for more details) and a full body HOG-based human detector \cite{peopleHOGdalalTriggs}. Detected objects are tracked simultaneously with a compressive tracker and a Kalman filter.} 

Munaro \etal \cite{munaro_tracking_2012,Munaro2014} 
find the optimal assignment of detections to tracks
in a Global Nearest Neighbor framework. Their matching likelihoods are obtained from the distance to the predicted position
and velocity, the probability of being a human as evaluated by a HOG based
human detector, and the similarity to the appearance model of the track. The latter is provided by an online Adaboost classifier trained on previous observations, and selects features in the color histogram space.
{Harville \etal \cite{Harville2004127} detect 
moving candidates by applying the background subtraction algorithm presented in \cite{harville_foreground_2001} to RGB-D data. The detected foreground objects are projected to a 2D reference plane where occupancy and height maps are generated. A box filter system is applied to the occupancy map such that 3D clusters not corresponding to a volume occupied by an average adult are filtered out. Their tracking Kalman filter state includes position in the reference plane and the height and occupancy maps data. These features are linearly combined to calculate the matching score that it is used in the measurements and update phases of the Kalman filter.}

\blue{Ma \etal \cite{DPMMultiple} present a tracking approach where a set of HOG based DPM detectors \cite{lsvm-pami} is applied to both depth and color images to detect body parts to enable their system to deal with a person's articulated motion. The conditional random field based approach of \cite{milanTracker} is used and extended to solve data association and trajectory estimation. In particular, person locations are inferred by minimizing an objective function, that includes detection matching, spatial correlation, mutual exclusion, temporal consistence, and regularization constraints. One interesting aspect of this method is that it can deal with flexible number and type of detectors.}

\subsection{Implementations of the variation of the pipeline}
\label{sub:coupled_tracking}

The detection of humans driven by generic full body descriptions, such as those mentioned earlier in Subsection  \ref{sub:sequential_det_track},
may sometimes be problematic, \eg when there is partial occlusion which can significantly alter the appearance of the target. 
In Section \ref{sec:characterization} (also see Fig. \ref{fig:pipelines}), we stated that in a variation to the common pipeline, some works attempt to address such difficult detections by exploiting trajectories and their representations in a combined detection and matching stage to enable more robust detection. {The trajectory {representations} provide descriptions of the targets, including first order motion models that enable} predictive tracking. 

After an 
ROI selection stage, Almaz\`{a}n and Jones \cite{almazan_tracking_2013} use {the} Mean-shift algorithm to find the ROI that best matches the 
appearance model of a target. For each trajectory, this search is initialised at the position predicted by a Kalman filter, and it is performed in the area defined by both the position variance estimated by the filter and by the ROI selection. The appearance model is made up of the color histogram of the highest 3D point of the person's cluster for each ground coordinate, and it is updated dynamically with each new observation as the weighted mean of the model at the previous frame and of the new histogram. The trajectory remains active until a number of frames after the target leaves the scene to allow its re-identification in case of temporary occlusion.

All other methods we review here that employ this variation of the MHT pipeline implement their first order motion model in a particle filter framework. A potential drawback of this is that particle filters
tend to be computationally expensive and may require optimisations to achieve practical running times.
In the works of Mu\~{n}oz-Salinas and co-workers \cite{MunSal2008,
Munoz2008b, munozsalinaspeople2009} {one particle filter is used per track, using a constant speed model to predict the next location of the target, and new target observations are searched for by maximizing
a detection probability.}
In \cite{MunSal2008,munozsalinaspeople2009},
candidates are identified in ROIs based on depth information, wherein
the probability of the presence of (any) person is computed based on heuristic rules on the number of points in a cluster and its maximal
height. In order to compute the probability of detecting the tracked person, this human presence probability is combined with an interaction factor that allows handling trajectory crossings by imposing a minimal separation between the positions of different people.
In \cite{MunSal2008}, the detection probability also includes the Bhattacharyya appearance similarity measure,
while in \cite{munozsalinaspeople2009} it uses a measure of confidence on depth. Hence, the trajectory representation in \cite{munozsalinaspeople2009} does not include any appearance model, and in \cite{MunSal2008} it models appearance by the color histogram of the cluster. This model is updated with new observations that have high detection and matching confidence
by the linear combination of the previous model and of the new histogram. This confidence condition avoids the model being updated when the detection contains parts of a different person, in case of close interaction between people.
In \cite{Munoz2008b}, 
the detection probability is made of three terms. It includes the probability of being a (facing) human, firstly by verifying that the cluster may be approximated by a vertical plane at the expected distance from the camera, and secondly, by evaluating the fitting of an ellipse
on the RGB image in order to validate the presence of the elliptical shape of a head
at this position. It also uses the Bhattacharyya appearance similarity measure to compare to
the trajectory representation's appearance model, made up of two (color) histograms inside two ellipses of pre-defined sizes and
respective positions that represent the head and torso respectively. This appearance model is updated dynamically as in \cite{MunSal2008}.

In all three methods in \cite{MunSal2008,
Munoz2008b, munozsalinaspeople2009}, new tracks are initialised when unknown targets are detected based on the use of generic person descriptions. 
In \cite{MunSal2008},
heuristic rules on the size and height of clusters are used. In
\cite{munozsalinaspeople2009}, confidence on depth is {added and new trajectories are initialised only after a few consecutive detections}.
In \cite{Munoz2008b}, the detection of new people is first performed by an Adaboost classifier trained on RGB images to detect {upper-bodies which} are {verified by heuristics} on their width and planarity using depth information. {Tracks are kept for a number of frames after occlusion or departure}.

{In \cite{Migniot2014}, Migniot \etal use a top-down view of a depth camera and propose a 2D model composed of two ellipsoids corresponding to head and shoulder regions which are obtained by simply thresholding the depth data. The chamfer distance between the observed regions and the ellipsoidal models is then used to assign the particle weights in their particle filtering tracker. In case of multiple persons in the scene, an independent tracker is created for each target.}

Choi \etal \cite{choi_detecting_2011,pamiSavarese} use particle-filtering with {Reversible} Jump Markov Chain Monte Carlo (RJ-MCMC) sampling to track multiple people simultaneously, as well as static non-human objects (obstacles) and the camera's position. Given the positions and velocities of all tracked targets and the results from generic person detectors applied to ROIs, at each iteration a move is attempted to initialise, delete or update a trajectory. The moves are sampled from the space of possible moves, one at a time, and the likelihood of the modified solution is estimated. Moves are accepted or rejected similar to MCMC sampling until the chain converges.
The moves are guided by the probability of
continuous tracking, based on a smooth target's motion constraint, that may also account
for people interactions \cite{pamiSavarese},
and the probability of being a human, as computed by a combination of HOG-based human detection,
face and motion detection, skin color and 2D shape recognition.
While \cite{pamiSavarese} accounts for the person's appearance in the likelihood, by computing the distance from a target-specific appearance-based mean-shift tracker \cite{Comaniciu2002}
that uses color information, \cite{choi_detecting_2011} does not use any appearance model. The appearance model for the tracker in \cite{pamiSavarese} is static though and built from a small number of consecutive frames,  in order to minimise tracking drifts.

{In the pedestrian tracking system presented by Gao \etal \cite{gaoIEEEITS}, a layered graph model is used to estimate pedestrian trajectories in RGB-D sequences. The color-based classifier of \cite{lsvm-pami} is used to detect target candidate regions from which several features such as 3-D position, appearance, and motion are extracted. The layered graph nodes represent the detected regions, and the edges the feature similarity. By minimizing the cost function of the graph, using an heuristic searching algorithm, the pedestrian trajectories obtained.}

\section{Survey by Use of Depth Information}
\label{sec:depth_use}

The works that we review in this paper seek to improve the stages of ROIs selection, human detection, and matching by the use of depth information as an additional cue.  
In this section, we outline how the use of depth, in combination with {RGB} information, can improve each and every stage of MHT.

\subsection{Use of Depth in ROI selection}
\label{sub:ROI}
Amongst the reviewed methods, all those that select ROIs rely heavily on depth and the additional information it provides on the scene geometry to identify the areas where people may potentially be found. As illustrated in the left branch of Fig.~\ref{fig:categories_depth}, we distinguish three categories of depth-based ROI selection methods, i.e. those based on the  estimation of the ground plane, those that model the scene's background, and those that detect motion. We now look at these categories in turn, and characterize the reviewed methods based on their ROI selection method - see columns 2 to 4 of Table \ref{tab:uses_depth}.

\textbf{Use of ground/ceiling plane --}
The assumption that people are usually located in areas of limited height
above the ground plane can greatly reduce the search space for detection, and this strategy has been used in many of the reviewed works. Munaro \etal \cite{Munaro2014} estimate the ground plane using a Hough-based method  \cite{Chambers2012}, and select as ROI the volume above the ground plane at typical human height. 

\begin{table*}[!ht]
\caption{Characterization of the reviewed methods based on their use of depth information}
\label{tab:uses_depth}
\vspace{0.2cm}
\centering
\resizebox{1\textwidth}{!}{
\begin{tabular}{>{\centering}p{4.8cm}|>{\centering}p{1.3cm}|>{\centering}p{1.9cm}|>{\centering}p{1.4cm}|>{\centering}p{1.4cm}|>{\centering}p{1.5cm}|>{\centering}p{1.5cm}|>{\centering}p{1.5cm}|>{\centering}p{1.7cm}}
\multirow{2}{*}{\bf Method} & \multicolumn{3}{c|}{\bf ROI Selection} & \multicolumn{3}{c|}{\bf Human Detection} & \multicolumn{2}{c}{\bf Matching}\tabularnewline
\cline{2-9} 
 & Ground Plane & Background subtraction & Motion Detection & 3D geometry & 2D depth classifier & 3D depth classifier & 3D Tracking & Joint RGB-D

description\tabularnewline
\hline \hline 
Bansal \etal \cite{Bansal2010} & \checkmark &  &  & & & \checkmark & &  \tabularnewline \hhline{-||---||---||--}
Salas and Tomasi \cite{salas_people_2011} &  & \checkmark &  & & & & \checkmark &  \tabularnewline \hhline{-||---||---||--}
Dan \etal \cite{Dan2012} &  &  &  & \checkmark & & & \checkmark & \checkmark \tabularnewline \hhline{-||---||---||--}
{Darrell \etal \cite{darrellIJCV}}&  & &  & \checkmark &  & &  &  \tabularnewline \hhline{-||---||---||--}
Han \etal \cite{han_employing_2012} & & & \checkmark & \checkmark & & & & \checkmark  \tabularnewline \hhline{-||---||---||--}
{Bajracharya \etal \cite{Bajracharya2009}} & {\checkmark} & & & {\checkmark} & & {\checkmark} & {\checkmark} & \tabularnewline \hhline{-||---||---||--}
Zhang \etal \cite{zhang_real-time_2013} & \checkmark &  & & \checkmark & & & \checkmark &  \tabularnewline \hhline{-||---||---||--}
{Galanakis \etal \cite{Galanakis2014}} & {\checkmark} & {\checkmark} & & {\checkmark} & & & {\checkmark} & {\checkmark} \tabularnewline \hhline{-||---||---||--}
Liu \etal \cite{icipColorHeightHist,Liu201516,LiuNEWDETECTOR} & \checkmark &  &  & \blue{\checkmark only in \cite{LiuNEWDETECTOR}}& & \checkmark & \checkmark & \checkmark \tabularnewline \hhline{-||---||---||--}
Luber \etal \cite{luberIROS11} and   &  &  &  & &  & & &  \tabularnewline  
{Linder \etal \cite{linderFUSION14}} &  &  &  & & \checkmark & & \checkmark & \checkmark  \tabularnewline \hhline{-||---||---||--}
{Ess \etal \cite{Ess2009}} & {\checkmark} & & & {\checkmark} & & & {\checkmark} &  \tabularnewline \hhline{-||---||---||--}
Jafari \etal \cite{aachenTracker} & \checkmark &  &  & & \checkmark & & \checkmark &   \tabularnewline \hhline{-||---||---||--}
Mu\~{n}oz-Salinas \etal \cite{Munoz2007} &  & \checkmark &  & & & & \checkmark &  \tabularnewline \hhline{-||---||---||--}
Munaro \etal \cite{munaro_tracking_2012,Munaro2014} & \checkmark &  &  & \checkmark & \checkmark & & \checkmark &   \tabularnewline \hhline{-||---||---||--}
{Almaz\`{a}n and Jones \cite{Almazan2014}} & & {\checkmark} & & {\checkmark} & & & {\checkmark} & {\checkmark} \tabularnewline \hhline{-||---||---||--}
{Bahadori \etal \cite{bahadori2005}} & \checkmark & \checkmark & \checkmark & \checkmark & & & \checkmark & \checkmark \tabularnewline \hhline{-||---||---||--}
{Beymer \etal \cite{beymer1999}} &  & \checkmark &  & \checkmark& \checkmark & & \checkmark &  \tabularnewline \hhline{-||---||---||--}
{Satake \etal \cite{satake2013}} &  &  &  &  & \checkmark &  & \checkmark &  \tabularnewline \hhline{-||---||---||--}
{Vo \etal \cite{Vo2014}} &  & \checkmark &  & \checkmark &  &  &  &  \tabularnewline  \hhline{-||---||---||--}
{Harville \etal \cite{Harville2004127}} & \checkmark & \checkmark &  & \checkmark &  & \checkmark & \checkmark &  \tabularnewline \hhline{-||---||---||--}
Almaz\`{a}n and Jones \cite{almazan_tracking_2013} &  & \checkmark & & \checkmark & & & \checkmark &   \tabularnewline \hhline{-||---||---||--}
Mu\~{n}oz-Salinas \etal \cite{MunSal2008,munozsalinaspeople2009} &  & \checkmark &  & \checkmark & & & \checkmark &    \tabularnewline \hhline{-||---||---||--}
Mu\~{n}oz-Salinas \etal \cite{Munoz2008b} &  &  &  & \checkmark & & & \checkmark & \checkmark  \tabularnewline \hhline{-||---||---||--}
Choi \etal \cite{choi_detecting_2011,pamiSavarese} & \checkmark & & \checkmark & & \checkmark & & \checkmark &  \tabularnewline \hhline{-||---||---||--}
{Migniot \etal \cite{Migniot2014}} &  &  &  & \checkmark & \checkmark & &  &  \tabularnewline \hhline{-||---||---||--}
{Gao \etal \cite{gaoIEEEITS}} &  & &  & \checkmark &  & & \checkmark &  \tabularnewline \hline
\blue{Ma \etal \cite{DPMMultiple}} &  & &  & \checkmark & \checkmark & & \checkmark & \checkmark \tabularnewline \hline
\end{tabular}
}
\end{table*}

\normalsize

Liu \etal \cite{icipColorHeightHist,Liu201516} detect 3D points that are local height maxima, located at a reasonable distance from the ground. ROIs are defined as vertical cylinders of fixed size centered on these maxima. \blue{In \cite{LiuNEWDETECTOR}, the same authors filter these positions by using a fast approach that applies typical head sizes and geometry to remove false candidates.}

{Ess \etal \cite{Ess2009} estimate the ground plane jointly with object detection in a Bayesian network. The ground plane is inferred from the bounding boxes of detected objects and the depth-weighted median residual between the ground plane estimate in the previous frame and the
lower regions of the depth image. Jafari \etal \cite{aachenTracker}} {first produce a rough estimation of the ground plane based on the known height of the camera, and then they project onto this plane the points that have a relative height of no more than 2m. 3D points that project in dense areas of the initial plane are excluded, and the remaining points are used to fit a more accurate plane to the ground surface} using RANSAC \cite{Fischler1981}. They then classify the remaining points into 3 different classes (`object', `free space', and `fixed structure' that usually denotes walls) based on their height and on their density when projected onto the ground plane. ROIs are searched for amongst the points labelled as `object', by clustering them based on the connectivity of their ground projections and by retaining the clusters that contain a high enough number of points. They are then divided into sub-clusters that are likely to contain single humans, using the Quick-Shift algorithm {\cite{Vedaldi2008}} that groups points around local maxima in the density of their ground projections. Similarly, Bansal \etal \cite{Bansal2010} also classify the 3D points into `object', `ground', and `overhanging structure' (\eg walls) using the distribution of heights in the cells of a grid superposed on the ground plane. The associations between these distributions and cells' labels are learnt off-line by kernel density estimation. Finally, a smoothing is applied to the pixels' labelling using a Markov Random Field  that penalizes neighboring pixels that have different labels. Pixels labelled as `object' are used in the detection phase to validate candidate detections.

Detecting and removing the ground plane from a point cloud also facilitates the clustering of the remaining points into separate objects, since they are no longer connected to each other through the floor. {Bajracharya \etal \cite{Bajracharya2009} project all 3D points onto a ground plane, presumably estimated based on known camera height and orientation. The resulting map is used to select areas of high density as ROIs of foreground points.}
Zhang \etal \cite{zhang_real-time_2013} exploit the known height of their camera to produce a rough estimation of the ground and ceiling planes, similarly to \cite{aachenTracker}. Then, at each new frame, they use the previously estimated planes to select 3D points within a distance threshold to the planes, that are used in a RANSAC algorithm to refine the planes' estimations. {After} removing the ground and ceiling planes, the remaining points are clustered, first by isolating regions of similar depths around local maxima in the depth distribution, and then, for each region, by extracting connected components in the image plane.
Munaro \etal \cite{munaro_tracking_2012} estimate the ground plane using a RANSAC-based least square method that is updated at each new frame to compensate for possible movements of the camera. The authors do not provide a detailed description of their RANSAC-based plane fitting stage. 
After suppressing the ground plane from the point cloud, they cluster the remaining points from their Euclidean distances. In order to avoid over-segmentation of objects, the neighbor clusters in ground plane coordinates are merged. {Humans} belonging to the same cluster are separated later in the detection stage, as will be explained in Section \ref{sub:human_detection}. 

Choi \etal use a similar strategy in \cite{choi_detecting_2011,pamiSavarese} {for} detection. After ground plane removal, they cluster 3D points 
and then select the clusters whose heights are within an acceptable range. HOG-based {detectors of both the upper and full body,} and {a} face detector are then applied to the clusters to generate their weak, initial detection hypotheses. In \cite{pamiSavarese}, a skin color detector, a motion detector, and a detector based on upper body shape from depth are also used. Galanakis \etal \cite{Galanakis2014} estimate the ground plane (without stating how) to discard any ROI points obtained by background subtraction that would be located on or close to the ground. 

{Bahadori \etal \cite{bahadori2005} apply off-line calibration to map their fixed stereo camera disparity data to the 3D world coordinate system. Their resulting  reference plane is used to track moving objects by using 3D spatial information. A similar calibration is applied also in the stereo system in \cite{Harville2004127}.} 

{\textbf{Background subtraction --}}
While similar to ground plane removal, the background subtraction strategy has the advantage of producing ROIs that are more likely to contain humans and to exclude static objects. Its drawback is that it requires learning a model of the background, and updating it in case of moving cameras or variable background. In the latter case, people need to be moving in the scene faster than the background model is updated, to be detected as moving objects. A background model that employs depth may be more robust than a color one to modifications of appearance that are not correlated with changes of the scene's geometry, such as due to illumination variations \cite{shottonReview,Camplani2014122}.

In \cite{almazan_tracking_2013,Almazan2014}, Almaz\`{a}n and Jones initialize a background model from the first few frames of a sequence. {The model is then updated progressively by a linear combination of the model's and current depth values where foreground objects are detected,} without modifying background areas that are assumed to remain unchanged. The result is that new background objects are eventually added to the model after they enter the scene, with the risk of adding stationary people when they stop moving for a significant amount of time. Foreground points are detected when the difference of their depth value with that of the model exceeds a threshold, {which was empirically established in \cite{almazan_tracking_2013}, and that accounts for the measured standard depth variation of the sensor as a function of the distance in \cite{Almazan2014}}. Foreground points are then projected onto a coarse horizontal grid, whose cells which contain a high enough number of foreground points are selected as ROIs.
Galanakis \etal \cite{Galanakis2014} also detect foreground points based on their difference with the depth values of a background model. No information is provided on the creation and possible update of the model. In a multi-camera setup, a global 3D coordinate system is used, and foreground points are reconstructed using triangular meshes. Triangles that are too close to the estimated ground plane are discarded. A top-down view of this scene - which in effect is a projection onto the floor - is generated using a GPU {and} used as a 2D map of the ROI {clusters}.

Mu\~{n}oz-Salinas \etal define a background model in \cite{Munoz2007} as a height map, \ie the map of maximum height for each ground plane coordinate. This model is built as the median of 10 consecutive maps, and it is updated every 10s. This update rate is chosen empirically based on the observed people's dynamics to reduce the risk of introducing a person who is temporarily standing in the scene into the model. Background subtraction is performed by selecting the points whose height are above the model's value. {Foreground} points are clustered using their projection on the ground plane, and the clusters that occupy a suitably large area and that contain enough points are selected as ROIs. These ROIs are used as human detections for the matching stage. A color-based face detector initializes new tracks. {In \cite{Harville2004127}, Harville applies the mixture of Gaussian based foreground segmentation method of \cite{harville_foreground_2001} to their stereo-based RGB-D data. The foreground objects are then projected to a 2D reference plane where occupancy and height maps are generated. These features are then used to track the foreground blobs with Kalman filters in the 2D reference plane.} 

Salas and Tomasi \cite{salas_people_2011} use the background model introduced by Gordon \etal \cite{Gordon1999} to combine color and depth in a 4D Gaussian mixture model. Foreground points are detected as those that are more than 3$\sigma$ away from the nearest background mode, and large clusters of 3D connected components are selected as ROIs. These ROI clusters are validated as humans or rejected in the detection stage by a  color-based HOG detector.
Mu\~{n}oz-Salinas and co-workers in \cite{MunSal2008,munozsalinaspeople2009} use a similar model that was defined in \cite{harville_foreground_2001}, which is updated by excluding points that belong to detected people. The foreground points are projected onto the ground plane for use in the detection and matching stages, and regions around local density maxima in this plane are selected as ROIs. {Bahadori \etal \cite{bahadori2005} propose a very simple unimodal background model by exploiting both intensity images and the estimated stereo disparity. The background model is dynamically adapted after a short initial phase where moving objects are assumed to be not in the scene. Moving object blobs are obtained by subtracting the actual data from the background model and then projecting it to the reference plane to be tracked.}

{Beymer	\etal \cite{beymer1999} employ the stereo based background subtraction algorithm proposed in \cite{EvelandStereoBG}. The background model is initialized with an empty scene. The foreground regions are then segmented to extract dominant disparity layers, assuming that different people in the scene may be located at different distances from the camera. The obtained blobs are then processed by the person detection module.}
{Vo \etal \cite{Vo2014,VoFaceDetector} identify the background areas combining the navigation information of the moving robot and a depth-based occupancy grid. Background areas are excluded from the candidate search space, speeding up the next steps of their algorithm.}

{\textbf{Motion detection --}}
For indoor applications, it may be reasonable to assume that moving objects are likely to be human, and to select areas with motion as ROIs for human detection. {Our previous discussion} on the respective sensitivities of depth and color to appearance changes for background subtraction also applies here, {and} motion may be detected more reliably using depth than from color only. Thus, authors such as Choi \etal  \cite{choi_detecting_2011,pamiSavarese} detect changes in 3D point clouds of consecutive frames, represented in octrees, following the method proposed in \cite{Kammerl2011}. The motion term in their estimation of human presence likelihood is then the ratio of moving pixels in the candidate region. 

Han \etal \cite{han_employing_2012} apply the same foreground detection technique as \cite{almazan_tracking_2013,Almazan2014}, but they use the previous frame as the background model. Thus, their foreground points selection is equivalent to selecting moving objects between two successive frames. The moving points are then clustered into ROIs based on the continuity of their values in the depth image.

\subsection{Human detection}
\label{sub:human_detection}

Depth information has been found by many authors to provide cues for human detection that are complementary to color-based appearance information. These cues mostly describe the 3D shape of the target, and they can be taken advantage of by (a) direct comparison against simple geometrical characteristics of a human shape, or (b) through the classification of 2D and 3D features, as detailed next. {Columns 5 to 7 of Table \ref{tab:uses_depth} summarize the reviewed methods based on their exploitation of these depth cues for people detection.}

\textbf{3D geometrical properties --} To speed up the detection process, many authors apply a cascade of small detectors to the ROIs, starting from the most lightweight ones, followed by the more computationally intensive ones on the few remaining candidates not dealt with by the earlier stages. A very fast and popular early detection stage is the assessment of the ROI clusters against simple geometrical constraints that are determined empirically. {In \cite{Bajracharya2009}, Bajracharya \etal select ROI clusters based on the expected width, height, and depth variance of a standing adult. Then a classifier on 3D features further refines the selection of clusters that have a human-like shape.}
In \cite{zhang_real-time_2013}, Zhang \etal first verify that the height of objects, as well as the number of points in their clusters, are within the expected ranges for a human target. Then, a random selection of normals to the cluster's surface vote to discard vertical (\eg wall) and horizontal (\eg tables) surfaces. Finally, a HOG and SVM-based detector is used to validate the remaining human candidates using RGB data.
In \cite{munaro_tracking_2012,Munaro2014}, Munaro \etal consider that ROI clusters may contain several {humans}, or a miscellany of {humans} and background objects. They extract sub-clusters that are likely to contain individual humans by detecting heads, denoted as local height maxima that follow heuristic rules on their distance from the scene floor and {on the minimal separation with others}. These initial detections are then validated or rejected as humans by a HOG-based detector on RGB data. {Vo \etal in \cite{Vo2014,VoFaceDetector} apply different geometric constraints to limit the search space of their skin and face detector modules. In particular, features like size and height (considering sitting and standing person) are used.}

In \cite{Munoz2008b}, Mu\~{n}oz-Salinas \etal detect the upper body using an Adaboost classifier with Haar-like features in color images, and then confirm these detections by verifying their width and planarity. For each positive classification, a binary mask of the upper body shape is applied to the depth image in order to compute the mean and standard deviation of the depth inside the template shape. These are used to estimate the probability for human detection, following heuristic assumptions on the expected width and planarity of a person. When confirmed by this test on depth distribution, the new detections are used to initialize new tracks.
Ess \etal \cite{Ess2009} combine the output of a (or any) color based person detector with depth based cues in a Bayesian network, where the object detection probability depends on the probability provided by the color based human detector, the probability of human presence given the scene geometry (using ROIs), and the probability of the detected object to be a human given its 3D geometrical properties, evaluated based on typical human height. The detection is also refined around the estimated location of the color detector by imposing a uniform depth inside its bounding box, when depth is sufficiently available.

{Other works, such as \cite{han_employing_2012,Dan2012,Migniot2014} limit their human detection stage to an assessment of the geometry of ROI clusters to achieve an even higher frame rate.} In \cite{Dan2012}, Dan \etal detect humans in top-down depth views by selecting local height maxima within a specified range that show the characteristic empirically-determined shape and size of head and shoulders when seen from above. {A similar top-down camera approach has been used by Migniot \etal \cite{Migniot2014} where the head and shoulder area is obtained by thresholding the depth data and fitting two ellipsoids to the identified regions. This model is then used to estimate body and head orientation.}  

Han \etal  \cite{han_employing_2012} only evaluate the height of moving ROI clusters, assuming that human-sized objects that move in an indoor 
environment are likely to be humans. Thus, ROI clusters, that have a height within a specified range that does not change significantly over five frames, are selected as human detections. {A similar approach is also presented in \cite{Harville2004127,bahadori2005,satake2013} where the height of the detected 3D blobs is used to discard moving objects that are unlikely to be people.}
Similarly, Almaz\`{a}n and Jones  \cite{almazan_tracking_2013,Almazan2014} select ROI clusters of moving points that have a high density 
when projected onto the ground. In \cite{almazan_tracking_2013}, detections are defined as areas of a pre-defined size around local density maxima in the ground projection map of the ROI points. In \cite{Almazan2014}, a blob detection technique is used, with smoothing and hole filling of the projected points into blobs, as well as filtering out those blobs that have a projected points density below a certain threshold. The authors note that the depth resolution of their sensors decreases with distance, producing an increasing {spread} of measured depth values around the exact values, i.e. a stretching of blobs on the line-of-sight of the camera.
Thus, the blobs are first normalized 
in a polar coordinate system. The density threshold is chosen as a function of depth, in order to compensate for the perspective effect that decreases the number of points in the blobs with distance.
Galanakis \etal \cite{Galanakis2014} also select blobs in their top-down 2D view of ROIs. Morphological operations are applied to a binary mask of their 2D view, and blobs that are too small are discarded.

In \cite{MunSal2008,munozsalinaspeople2009}, Mu\~{n}oz-Salinas and co-workers compute the likelihood of human presence in candidate regions (\ie regions around local density maxima of foreground points projected onto the ground plane), based on the maximal height and the number of points in these regions that follow empirically established expected values and associated standard variations. This likelihood is used both for detecting new people to initialize tracks and for tracking existing targets. 

{Darrell \etal \cite{darrellIJCV} segment the disparity stereo map  with a simple combination of a gradient operator and thresholding based on the typical volume occupied by a person facing the camera. Large connected components are then considered as possible human candidates, and head locations are estimated at the top of each connected component. A combination of face and skin detectors is used to rule out false detections.}

\blue{Liu \etal \cite{LiuNEWDETECTOR}, improved the detection performance of their previous algorithms \cite{icipColorHeightHist,Liu201516} both in terms of accuracy and processing speed by using a cascade of classifiers based on 3D geometry data. They use a very fast filter based on empirical thresholds on the typical human head size. A second classification stage, based on a Ring-wedge Mask Detector \cite{Ganotra2004167}, is then applied to identify the head and shoulders region.}

\blue{Ma \etal \cite{DPMMultiple} use a pool of classifiers based on color and depth data to detect the human body and its articulated parts. The 3D spatial structure of the tracked object's parts is taken into account such that the detector pool learns pre-determined configurations, and hence the system is able to cope with pose variations.}

\textbf{Classifier of 2D features in the depth map --}
Classifiers that are traditionally used on grey-level or color  images may also be applied to depth data, or disparity maps in stereo imaging, to recognize the 2D shape of a human. Munaro \etal  \cite{Munaro2014} apply a Haar-like feature classifier in a cascade to both the color image and the disparity map, to exploit different and independent features that increase the detection rate and reduce the number of false positives. Luber \etal \cite{luberIROS11} introduce a variant of the HOG detector for depth maps, the Histogram of Oriented Depth (HOD), that they use in a SVM classifier to compute probabilities which are linearly combined with the ones obtained from a classical HOG-based SVM classifier on RGB data to detect humans. \blue{A similar approach is used in \cite{DPMMultiple} where several DPM classifiers \cite{lsvm-pami} trained on HOG features extracted from the depth maps are used in the detection phase.} 

Template matching in depth images has also been used \cite{beymer1999, satake2013,choi_detecting_2011,pamiSavarese,aachenTracker} for recognizing the 2D shape of the upper-body. Choi \etal \cite{choi_detecting_2011,pamiSavarese} compute the likelihood of the depth image to contain a person by template matching of the thresholded depth map with the upper body shape. This probability is combined with the output of a HOG-based detector, and of face, skin color, and motion detectors, to obtain the human presence likelihood term in their tracking algorithm. 
Similarly, Jafari \etal \cite{aachenTracker} perform template matching of the depth map with a depth template of the upper body. This depth-based detection is used in close-range images, while a color-based HOG detector allows for detecting people at a further range where depth sensors may not operate satisfactorily. {In \cite{beymer1999}, the binary template is applied in a classic fashion to foreground blobs and a person is detected when the response is above a certain threshold. 
Satake \etal \cite{satake2013} apply a set of three binary templates \cite{satake2009robust}, containing frontal and side views of head and shoulders, to the disparity map. The sum of squared distance criterion is then used to select human candidates. Detections are checked by using a SVM classifier trained on HOG features.}

\textbf{Classifiers on the 3D point cloud --}
Similarly to \cite{choi_detecting_2011,pamiSavarese,aachenTracker}, template matching of a human shape may also be performed in 3D. Bansal \etal \cite{Bansal2010} adapt a 3D template to the camera view-point, before its correlation with the disparity map is computed for each ground plane coordinate. Local maxima in this correlation map, together with neighboring correlation values above 60\% of the associated maximum, are selected as initial detection candidates. These regions are refined by discarding points with divergent depths, and by selecting areas with a high density of edges in the color image. 

Bajracharya \etal \cite{Bajracharya2009} apply a linear classifier to a number of features derived from the 3D points of detected candidates. Some of these features capture the {variance} of the height of the points  within the candidate, and the object's size and extent. {Three} rotationally invariant features also account for the eigenvalues of the point cloud's covariance matrix.

In order to avoid making hard assumptions on the shape of a human body or upper body, Liu \etal \cite{icipColorHeightHist,Liu201516} train an SVM classifier on two features computed from the height and color distributions of 3D points. Their features are a histogram of the heights of the upper body, and a joint color and height histogram of the head, respectively. The upper body and head points are found in regions of pre-defined sizes in the ROI clusters. 
{Harville \etal \cite{Harville2004127} apply a box filter, set by considering the average adult human height and  torso width, to the occupancy map corresponding to the segmented 3D foreground clusters.  The peak of the response is thresholded to discard false positive detections.}

\subsection{Use of depth in matching}
\label{sub:identification}

This section reviews how the use of depth information reduces ambiguities for establishing correspondences of detected people against existing tracks through (a) the provision of 3D trajectories, and (b) by enhancing description of the target in combination with color. These two uses of depth information for matching in the reviewed methods are summarized in the last two columns of Table \ref{tab:uses_depth}.

{\bf 3D tracking --}
The majority of methods reviewed in this survey construct trajectories in the 3D space to facilitate 3D tracking. This allows better handling of trajectory crossings when objects move past each other in the scene in the camera's viewpoint. We now highlight the role of depth position information in the matching stage which was described earlier in Section \ref{sec:pipelines}.

Dan et al. \cite{Dan2012} place their camera on the ceiling with a top-down view, therefore the 2D coordinate system of the image can be seen as a good approximation of the 2D ground plane coordinate system. Then, they match detected 3D shapes in adjacent frames from their degree of physical overlap.
Galanakis \etal \cite{Galanakis2014} also track people in a top-down view of the 3D scene {rendered from multiple views}, by comparing their distance to predicted target positions on the 2D ground plane. 

{Gao \etal \cite{gaoIEEEITS} employ depth data to build a 3D layered graph model of the scene to solve possible occlusions, and thus, they boost their proposed tracking algorithm.} 
In \cite{salas_people_2011}, Salas and Tomasi exploit 3D location information for computing one-to-one correspondences between candidates, by including a term based on their separating distance in their appearance and motion based correspondence likelihood formulation.
Similarly, the various authors of \cite{bahadori2005,beymer1999,satake2013,Bajracharya2009,zhang_real-time_2013,icipColorHeightHist,Liu201516,luberIROS11,aachenTracker,Ess2009,Munoz2007,munaro_tracking_2012,Munaro2014,Almazan2014} all perform matching by determining the 3D distance of a detected candidate to its predicted position. In \cite{Harville2004127,almazan_tracking_2013,MunSal2008,Munoz2008b,munozsalinaspeople2009,choi_detecting_2011,pamiSavarese,DPMMultiple}, 3D position predictions are used to initialize the search for targets in the 3D space.

Only {three} works described in Section \ref{sec:pipelines} do not exploit the 3D trajectory information. In \cite{han_employing_2012}, Han \etal use similarities in color, and variations of depth across two adjacent frames, in order to compute matching correspondences, without taking into account the 3D coordinates. In \cite{Bansal2010}, although Bansal \etal use 3D coordinates for ROI selection for their human detection stage, and for camera motion estimation, their matching stage is performed in 2D. {Similarly, Vo \etal \cite{Vo2014} implement their compressive tracking and Kalman filter by considering the target's movements only in the image plane.} 

{\bf Joint RGB and depth description --}
The fusion of  color and depth information allows for more reliable correspondence of detected candidates to tracks. An example of such fusion   is in Luber \etal \cite{luberIROS11}, where they build their model from a  combination of three possible features: Haar-like features in intensity and depth images, and \textit{Lab} color feature in the RGB image. Several such features are calculated from small rectangles, randomly sized and positioned inside the bounding box of the detected person. A combination of a few of these features is selected by on-line boosting to produce a classifier that attempts to distinguish the tracked person from its surroundings. 
Liu \etal \cite{icipColorHeightHist},\cite{Liu201516} use a joint color and height histogram of the full body as their appearance model. The likelihood of new detected candidates to match this model is computed using the Bhattacharyya similarity measure. Similarly, Almaz\`{a}n and Jones \cite{Almazan2014} model people's appearances from a height histogram associated with color distributions for each histogram bin, approximated by 3D Gaussians in the RGB space. 
Dan \etal \cite{Dan2012} assess the correspondence between two detected candidates by linearly combining a Bhattacharyya measure of similarity of their color histograms, and the overlap of the 3D shapes of both candidates. This last value, in addition to accounting for the distance in the 3D space between the candidates, may also capture the similarity of their shapes if their 3D locations are close enough.
{Beymer \etal \cite{beymer1999} use an intensity model and average disparity value to describe a person. Both are linearly updated, and their drift is limited by applying their person detector confidence as a smoothing factor.}

Han \etal \cite{han_employing_2012} propose the use of depth for generating an appearance model, where a silhouette obtained from depth information helps in isolating the relevant parts of the body from which a color-based appearance model is built. The neck and waist are identified as local width minima of the silhouette along the vertical direction. They divide the color image of the person into  head, torso, and legs areas. Torso and legs colors are then used to build the appearance model, by concatenating histograms of color and texture for both regions.
Galanakis \etal \cite{Galanakis2014} also exploit depth to produce a two-part color histogram model of upper and lower body, using their textured mesh representation obtained during their ROI selection stage. The mesh is divided into lower and upper body parts at an empirically determined height. 

In \cite{Munoz2008b}, Mu\~{n}oz-Salinas \etal assume planarity of standing people in order to compute a single-valued 
depth term of the RGB and a depth based likelihood of a target detection. The mean depth of a candidate region is assessed against a normal distribution with mean equal to the predicted target's distance to the camera, and standard deviation chosen heuristically and decreasing with an increased confidence in depth (measured as the proportion of pixels in the region that have a depth measure). Two depth terms are computed for the head and torso separately. The detection likelihood of a target also includes the comparison to two color histogram-based appearance models for the head and torso respectively, using the Bhattacharyya measure, and the assessment of the fitting of an ellipse on the color image at the expected head location, using image gradients.

\section{Considerations on the practical applications of MHT}
\label{sec:applications}

The methods presented in this survey are, almost always, customised for specific scenarios or application areas by employing assumptions on aspects, such as the position of the camera(s) (e.g. static or mobile, top-down or head-level view), the geometry of the scene, and {the} generic description of a person. In order to guide the reader in their choice of RGB-D MHT method, we next outline the conditions of use of the reviewed methods. These are also  summarized in Table  \ref{tab:practical_applications}.

\subsection{Type of depth sensor}

Historically, depth has been mostly obtained from passive stereo cameras, which offered a cheaper alternative to other technologies such as active sensor cameras. Depth from stereo vision is still widely used in MHT methods, such as \cite{beymer1999,darrellIJCV,Harville2004127,bahadori2005,satake2013,Munoz2007,MunSal2008,Munoz2008b,munozsalinaspeople2009,Bansal2010,aachenTracker,Ess2009,Bajracharya2009}. The recently introduced and affordable Kinect camera (and those like it) generate depth from structured light and are  more convenient to use than stereo vision for indoor scenes, since they do not require calibration and an elaborate computation of a disparity map. Thus, computer vision researchers are increasingly adopting such cheaper and more immediate technology for RGB-D MHT  when it can sufficiently serve their purpose, such as in \cite{Migniot2014,Vo2014,VoFaceDetector,luberIROS11,choi_detecting_2011,salas_people_2011,Dan2012,munaro_tracking_2012,han_employing_2012,pamiSavarese,zhang_real-time_2013,almazan_tracking_2013,Almazan2014,icipColorHeightHist,Liu201516,Munaro2014,aachenTracker,Galanakis2014,DPMMultiple}.

Most of the methodologies presented in this survey could use \blue{passive and active sensors} interchangeably, including those that extract features directly from disparity maps, as disparity and depth maps have similar properties. However, the optimal conditions of use for both types of sensors differ significantly, both in terms of depth range and illumination conditions. For example, the depth range of structured light cameras tends to be more limited than that of stereo vision, and they   are also more sensitive to infra-red light, which makes them unsuitable for outdoor uses. On the other hand, \blue{color based} stereo cameras require good illumination conditions and they may not operate in dark environments for example. Moreover, the additional processing time required to obtain disparity from stereo data can be critical for real time applications. These particularities were highlighted by Jafari et al., who used both sensors in \cite{aachenTracker} to track people in close and far ranges. {The second column of Table \ref{tab:practical_applications} shows the type of sensor used in the works in this survey.} 

\subsection{Camera Position}

{\bf Handling of moving cameras --}
Some applications rely on static sensors that provide a stable background model, \eg \cite{bahadori2005,Harville2004127,beymer1999,Munoz2007,MunSal2008,munozsalinaspeople2009,salas_people_2011,almazan_tracking_2013,Almazan2014,Galanakis2014}, especially when this model is static itself and not updated on-line to account for camera movements, as in \cite{salas_people_2011,almazan_tracking_2013,Almazan2014} and we presume in \cite{Galanakis2014}. Some methods attempt to update the background model continuously, \eg \cite{bahadori2005,Harville2004127,beymer1999,Munoz2007,MunSal2008,munozsalinaspeople2009}. Although these MHT methods did not present any experiments with mobile cameras, the authors of \cite{Munoz2007} state that their method has been developed with 'human-mobile robot' interaction in mind, and that their background modelling technique is specially appropriate for mobile devices. Indeed, these models may be able to adapt to camera motion, provided that the update rate is faster. The implementation of this strategy is not easy, since, as  discussed in Section \ref{sub:ROI}, an update rate that is too fast would be likely to result in slow people being included in the background model. Thus, as the authors explain, it has to be tuned depending on the application.

On the contrary, methods that do not assume a static or nearly static background can generally be used with a mobile camera, such as a PTZ or one mounted on a mobile robot. Some methods assume that both person and camera motion are smooth, and they treat their combined effects as that of a single speed for the tracked person relative to the camera \cite{Munoz2008b,munaro_tracking_2012,zhang_real-time_2013,satake2013}. 
Others exploit the global 3D coordinate system provided via the depth dimension in order to track the camera's movements. Choi \etal  \cite{choi_detecting_2011} {and Vo \etal \cite{Vo2014}} compute the position of the camera using the ROS library \cite{ROS} in order to project target locations onto the global 3D coordinate system, and Bansal \etal \cite{Bansal2010}, Bajracharya \etal \cite{Bajracharya2009}, Ess \etal \cite{Ess2009}, and Jafari \etal \cite{aachenTracker} do the same using visual odometry. {In \cite{Ess2009}, the visual odometry algorithm is improved by feedback from the tracker which helps  avoid using areas that are likely to contain moving objects.} In their later work, Choi \etal \cite{pamiSavarese} estimate both the motion of the camera and the humans in the scene in their combined detection and matching stage. \blue{In general, these approaches assume that the camera is moving, at least locally, on a mostly flat ground plane.}

\begin{table*}
\caption{Conditions of use of the presented methods
\label{tab:practical_applications}}
\centering
\scriptsize
\resizebox{1\textwidth}{!}{
\begin{tabular}{*{2}{>{\centering}m{0.08\linewidth}|}*{2}{>{\centering}m{0.065\linewidth}|}>{\centering}m{0.14\linewidth}|>{\centering}m{0.06\linewidth}|>{\centering}m{0.10\linewidth}|>{\centering}m{0.06\linewidth}|>{\centering}m{0.18\linewidth}}
{\bf Method} & {\bf Sensor type} & {\bf Handle a mobile camera} & {\bf Handle multiple cameras} & {\bf Camera location constraints} & {\bf Real-time} & {\bf \blue{Processing hardware}} & {\bf Require flat ground} & {\bf Other special requirements}\tabularnewline
\hline \hline
Bansal \etal \cite{Bansal2010} & Stereo & \checkmark & \text{x} & Roughly frontal view 
 & 10 fps & \blue{CPU Intel Dual-Core} & \checkmark & None identified \tabularnewline \hline
Salas and Tomasi \cite{salas_people_2011} & Structured light & \text{x} & {Not tested} & Roughly frontal view 
& No data & \blue{No data} & \checkmark & Limited to standing people 
\tabularnewline \hline
Dan \etal \cite{Dan2012} & Structured light & \text{x} & {Not tested} & Vertical top-down view & 55 fps on QVGA stream & \blue{CPU 2.4GHz, 4GB RAM} & \checkmark & Limited to standing adults 
\tabularnewline \hline
{Darrell \etal \cite{darrellIJCV}} & Stereo & Not tested & \text{x} & Roughly frontal view 
 & 12 fps & \blue{No data} & \text{x} & None identified \tabularnewline
 \hline
Han \etal \cite{han_employing_2012} & Structured light & {Not tested} & \text{x} & Roughly frontal view 
 & 10 fps (2 people)& \blue{CPU Dual core 2.53GHz, 4GB RAM} & \text{x} & Limited to standing adults -- People must be moving to be detected \tabularnewline \hline
{Bajracharya \etal \cite{Bajracharya2009}} & Stereo & {\checkmark} & {Not tested} & {None identified} & {5--10 fps}& \blue{No data} & {\text{x}} & {Limited to standing adults} 
\tabularnewline \hline
Zhang \etal \cite{zhang_real-time_2013} & Structured light & \checkmark & {Not tested} & Roughly frontal view 
 & 7--15 fps& \blue{CPU 2.0GHz, 4GB RAM} & \checkmark & None identified \tabularnewline \hline
{Galanakis \etal \cite{Galanakis2014}} & {Structured light} & {\text{x}} & {\checkmark} & {Multiple views 
from different angles are desirable} & Real-time (no exact data) & \blue{GPU NVIDIA GTX680} & {\text{x}} & {None identified} \tabularnewline \hline
Liu \etal \cite{icipColorHeightHist,Liu201516} & Structured light & {Not tested} & {Not tested} & None identified & 30-50 fps& \blue{CPU  i5-2500, 8GB RAM} & \checkmark & May be limited to standing adults 
\tabularnewline \hline
Luber \etal \cite{luberIROS11} & Structured light & {Not tested} & \checkmark & Roughly frontal view 
 & No data & \blue{No data} & \text{x} & May be limited to standing people 
 \tabularnewline \hline
{Ess \etal \cite{Ess2009}} & {Stereo} & {\checkmark} & {Not tested} & {Roughly frontal view when using HOG based detectors} & 3 fps & \blue{GPU nVidia GeForce 8800 and CPU 2.66 GHz}  & {\checkmark} & {None identified} \tabularnewline \hline
Jafari \etal \cite{aachenTracker} & Combined stereo and structured light & \checkmark & \checkmark & Roughly frontal view 
 & {18-24} fps & \blue{CPU i7-3630QM and GPU NVIDIA GeForce GT650m, 12GB RAM}
 & \checkmark & May be limited to standing people in the far range 
 \tabularnewline \hline
Mu\~{n}oz-Salinas \etal \cite{Munoz2007} & Stereo & {Not tested} & {Not tested} & Roughly frontal view 
 & 10 fps & \blue{CPU 3.2 GHz Pentium IV} & \checkmark & People only detected in a close range (by face detector) but tracked on a larger range \tabularnewline \hline
Munaro \etal \cite{munaro_tracking_2012,Munaro2014} & Structured light & \checkmark & \checkmark & Roughly frontal view 
 & 30 fps \cite{Munaro2014}, 26 fps \cite{munaro_tracking_2012} & \blue{CPU Xeon E31225 3.10GHz \cite{munaro_tracking_2012}} & \checkmark & Minimal separation between people's heads: 30 cm -- May be limited to standing adults 
 \tabularnewline \hline
Almaz\`{a}n and Jones \cite{almazan_tracking_2013,Almazan2014} & Structured light & \text{x} & \checkmark & None identified & No data & \blue{No data} & \text{x} & Stationary people may not be detected after a while \tabularnewline \hline
{Bahadori \etal \cite{bahadori2005}} & Stereo & \text{x} & {Not tested} & Fixed to ceiling, pointing down at $30^\circ$  
 & 10 fps & \blue{CPU 2.4 GHz} & \checkmark & May be limited for other camera configurations, 3D coordinate system calibration \tabularnewline  \hline
{Beymer \etal \cite{beymer1999}} & Stereo & \text{x} & Not tested & Parallel to the ground floor 
 & 10 fps & \blue{No data} & \text{x} & None identified \tabularnewline
 \hline
 {Satake \etal \cite{satake2013}} & Stereo & \checkmark & Not tested & None identified 
 & 9 fps & \blue{No data} & \text{x} & Developed for wheel-chair applications, camera placed around 1m height \tabularnewline
 \hline
{Vo \etal \cite{Vo2014,VoFaceDetector}} & Structured light & \checkmark & \text{x} & None identified 
 & 23 fps &\blue{CPU 2.4 GHz} & \text{x} & Developed for robot applications, camera placed around 1m height \tabularnewline
 \hline
{Harville \etal \cite{Harville2004127}} & Stereo & \text{x} & Not tested & None identified 
 & 8 fps&\blue{CPU 750 MHz} & \checkmark & None identified \tabularnewline
 \hline
Mu\~{n}oz-Salinas \etal \cite{MunSal2008,munozsalinaspeople2009} & Stereo & {Not tested} & \checkmark & None identified & 15~fps \cite{MunSal2008} and 100~fps \cite{munozsalinaspeople2009} (4 people)& \blue{AMD Turion 3200, 1GB of RAM} & \checkmark & None identified \tabularnewline \hline
Mu\~{n}oz-Salinas \etal \cite{Munoz2008b} & Stereo & \checkmark & {Not tested} & Frontal view at head level & 23 fps (3 people)&\blue{AMD-K7 2.4 GHz} & \text{x} & Limited to standing people \tabularnewline \hline
Choi \etal \cite{choi_detecting_2011,pamiSavarese} & Structured light & \checkmark & {Not tested} & Roughly frontal view 
 & 5-10 fps & \blue{GPU} & \text{x} & May be limited to adults 
\tabularnewline \hline
{Migniot \etal \cite{Migniot2014}} & Structured light & \text{x} & Not tested & Vertical top-down view 
 & 40 fps& \blue{CPU 3.1 GHz} & \text{x} & None identified \tabularnewline
 \hline
 \blue{Ma \etal \cite{DPMMultiple}} & \blue{Structured light} & \checkmark & \blue{Not Tested} & \blue{None identified} 
 & \blue{No data} & \blue{No data} & \blue{\text{x}} & \blue{Hand labelled data of body parts to train DPM} \tabularnewline \hline
\end{tabular}
}
\end{table*}

Note that the works in \cite{Munoz2007,icipColorHeightHist,Liu201516,MunSal2008,munozsalinaspeople2009,DPMMultiple}, although not tested with mobile cameras, perform tracking in a global 3D coordinate system similarly to \cite{Bansal2010,choi_detecting_2011,aachenTracker}, and we believe could therefore apply the same mobile camera-handling strategy if combined with camera motion estimation. \blue{These approaches can be successful in a moving camera scenario if the camera position requirements (discussed in the next section) can be generally met.} 

The works in \cite{luberIROS11,han_employing_2012,Munaro2014} also could employ the same `smooth relative-speed strategy' as \cite{Munoz2008b,munaro_tracking_2012,zhang_real-time_2013}. 
The possibility of using mobile cameras with the reviewed methods is indicated in the third column of Table \ref{tab:practical_applications}.

{\bf Handling of multiple cameras --}
The works in \cite{beymer1999,Harville2004127,satake2013,gaoIEEEITS,luberIROS11,munaro_tracking_2012,aachenTracker} can exploit {information} from multiple cameras simultaneously and fuse detections from independent cameras at the matching stage. This requires the relative positions and orientations of the cameras to be known or estimated off-line. In particular, \cite{munaro_tracking_2012} has been extended to the multi camera scenario in \cite{openPTrack,MunaroSWJournal}. This strategy would be accessible to all methods that apply the main MHT pipeline and perform tracking in a 3D global coordinate system.

This multi-sensor data fusion strategy is not possible in works that {apply the variation of the MHT pipeline since} they do not perform the detection and matching stages sequentially, such as in \cite{almazan_tracking_2013,choi_detecting_2011,pamiSavarese,MunSal2008,munozsalinaspeople2009,Munoz2008b}. However, in \cite{almazan_tracking_2013,Almazan2014,MunSal2008,
munozsalinaspeople2009}, detection and matching are performed on a global representation of data on the 2D ground plane, which is generated in \cite{almazan_tracking_2013,Almazan2014,munozsalinaspeople2009} from the point clouds of several cameras.
The methods in \cite{bahadori2005,choi_detecting_2011,pamiSavarese,Munoz2008b} use 2D color image based people detectors, but they track people in a 3D space. Thus, they could use multiple cameras if all the transformations from the individual image spaces to the global 3D space are known.
{Similarly to \cite{almazan_tracking_2013,Almazan2014,MunSal2008,
munozsalinaspeople2009}, Galanakis \etal \cite{Galanakis2014} detect and  then track people in a common 2D coordinate system.}
The fourth column in Table \ref{tab:practical_applications} indicates which of the reviewed works can (or could) handle multiple cameras.

{\bf Requirements on the camera's position and orientation --}
\label{subsec:applicationsREQ}
Methods that use human detectors that are trained on specific view positions and angles, such as HOG trained from roughly frontal views, require similar views of people. This is the case in \cite{munaro_tracking_2012,Munaro2014,luberIROS11,zhang_real-time_2013,Munoz2008b,salas_people_2011,Bansal2010,aachenTracker}, and also in the implementation of \cite{Ess2009}, although the authors stress that other color based detectors can be used. 
Similarly, the works in \cite{choi_detecting_2011,pamiSavarese,Munoz2007} employ face detectors, and require a roughly frontal view of the face to be visible in a significant number of frames. The methods in \cite{darrellIJCV,Munoz2007,Munoz2008b} were specifically designed for a camera located at head (or just under head) height. In particular, the work in \cite{Munoz2008b} assumes the human shape as seen by the camera can be approximated by a vertical plane. 
Han \etal \cite{han_employing_2012} also require a frontal view for analyzing human silhouettes, as explained earlier in Section \ref{sub:identification}. The methods in \cite{Dan2012,Migniot2014} operate on a top-down view due to their specific detection strategy centered around monitoring humans seen from above. {In \cite{Galanakis2014}, a sufficient coverage of the scene by multiple cameras at various viewing angles is preferred to produce 3D textured meshes of humans.}
{In \cite{bahadori2005}, the camera is placed on the ceiling at an angle of $30^\circ$, so as to reduce occlusions as upper body parts are always visible.} 
\blue{A similar camera position is used in the e-health application presented by Ma \etal in \cite{DPMMultiple}. However, their proposed system, based on different DPM classifiers, is able to deal with considerable variations of human body pose, hence ensuring also a certain invariability to camera viewpoints.} 
{Beymer \etal \cite{beymer1999} propose a 3D motion model based on the assumption that the stereo camera is placed parallel to the ground floor. In \cite{satake2013},  the system has been specifically designed for a wheel-chair navigation system, and the stereo camera is placed at an approximate height of 1m. }
The various requirements and limitations of the camera position and orientation are summarized in column 5 of Table \ref{tab:practical_applications}.

\subsection{Speed of computation}

The works in \cite{salas_people_2011,luberIROS11,almazan_tracking_2013,Almazan2014,DPMMultiple} provide no computational information. {Harville \etal \cite{Harville2004127} report a processing rate of 8fps, however this is obtained using obsolete hardware and it would dramatically improve if tested on current workstations.} The rest of the methods we review claim real-time performances, with the exception of Ess \etal \cite{Ess2009}, who report a running time of 300ms per frame on a GPU, plus an additional (off-line) 30s for the color based human detector. Their method can be used with other, more efficient, color based detectors.  

Running times \blue{and the hardware platforms used}, when available, are reported in column 6 \blue{and 7} of Table \ref{tab:practical_applications}.
Methods that use stereo vision suffer from the overhead of deriving disparity maps, while depth information is readily available {from} structured light-based {sensors}. Some authors, such as Bansal \etal in \cite{Bansal2010}, speed-up this computation using a GPU.

Works such as in \cite{han_employing_2012,MunSal2008,munozsalinaspeople2009}  report performances that vary significantly according to the number of people being tracked. This is particularly the case in methods that use multiple trackers for individual people, such as {the 3D Kalman filter in \cite{satake2013}} or particle filters in \cite{MunSal2008,munozsalinaspeople2009,Munoz2008b,Migniot2014}. Such methods also need to establish a trade-off between the number of particles used and the accuracy and robustness of tracking. 

Jafari \etal \cite{aachenTracker} exploit both depth and color information in complementary distance ranges, and speed-up the total process from 33fps (on GPU) to 18fps by applying the computationally intensive color detector only in far ranges (over 7m) where the depth-based detector does not operate.

\blue{Finally, Liu \etal \cite{Liu201516} report a processing rate range of 30-50 fps, without the use of GPU hardware, for their detection and tracking system. In addition, in their more recent work \cite{LiuNEWDETECTOR}, they boosted the detection phase by using a cascade of classifiers on top of their depth-color histogram model. This meant that their detection module can operate in a range of 77-140 fps, however no rate is given for the entire detection and tracking processing.}

\subsection{Specific Constraints}
\label{subsec:specialReq}

{\bf Flat ground --}
Methods that detect ROIs based on an estimation of the ground plane, as detailed in Section \ref{sub:ROI}, cannot handle environments where the ground cannot be approximated by a plane. This is the case, for example, of staircases, where Munaro \etal report worse results in \cite{munaro_tracking_2012}. Similarly, the methods in \cite{Bansal2010,aachenTracker} classify the scene into general categories that include a flat ground and vertical structures, and would most likely not generalize well to a staircase environment.

In both \cite{Dan2012} and \cite{Migniot2014}, people {are detected by thresholding} the distance of their head to the camera, which has to be within an acceptable range. Therefore, although there is no hard constraint on ground planarity, varying ground level can influence the head-to-camera distances significantly. 

Methods, such as those in \cite{almazan_tracking_2013,Almazan2014,icipColorHeightHist,Liu201516,Munoz2007,MunSal2008,munozsalinaspeople2009,Galanakis2014,Bajracharya2009}, project detections onto a flat ground plane. In \cite{almazan_tracking_2013,Almazan2014,Galanakis2014,Bajracharya2009}, ROIs are not selected based on height from this plane\footnote{{In \cite{Bajracharya2009}, ROI clusters are selected based on their absolute height, and in \cite{Galanakis2014}, only a few points close to the ground plane are discarded, not the full ROI clusters.}}, so the flat ground assumption does not need to correspond to reality. However, this is not the case in \cite{icipColorHeightHist,Liu201516,Munoz2007,MunSal2008,munozsalinaspeople2009} where people have to be located in a relatively narrow band above the floor to be detected. {In \cite{bahadori2005,Harville2004127} a reference plane is used to track 3D blobs in a real world coordinate system. As the camera and real world scene are calibrated, the reference plane does not have to be necessarily flat}. Column 7 in Table \ref{tab:practical_applications} indicates which of the methods operate only in ground plane scenarios.

{\bf Constraints on pose --}
Several works, \eg  \cite{beymer1999,Harville2004127,bahadori2005,Migniot2014,Vo2014,munaro_tracking_2012,Munaro2014,Dan2012,han_employing_2012,choi_detecting_2011,pamiSavarese,icipColorHeightHist,Liu201516,Munoz2007,MunSal2008,munozsalinaspeople2009,Bajracharya2009} select ROIs based on height and volume assumptions derived from a model of a standing adult person. Such methods may not be able to detect and track, e.g., children, adults with abnormal  heights, and sitting people, if the {acceptable ranges} for height and volume are not chosen appropriately. This is the case, for example, for Choi \etal  \cite{choi_detecting_2011} and Han \etal  \cite{han_employing_2012}, who filter heights in ranges of 1.3 to 2.3~m and 1.5 to 2~m respectively. Other authors, such as Zhang \etal \cite{zhang_real-time_2013} and Liu \etal \cite{icipColorHeightHist,Liu201516}, accept quite larger range of values (0.4 to 2.3~m and 0.6 to 2~m for height, respectively), to prepare their ROIs to handle children or people who may not be standing. {Ess \etal \cite{Ess2009} 
suggest the possibility of detecting children by increasing  the standard deviation of their normal height distribution.

Methods that use full-body detectors such as HOG and HOD, i.e. \cite{Vo2014,munaro_tracking_2012,Munaro2014,luberIROS11,salas_people_2011,
aachenTracker}, may also struggle to detect sitting people if these detectors are trained on standing people only. To alleviate this shortcoming, Choi \etal \cite{choi_detecting_2011,pamiSavarese} combine full-body and upper-body detectors, in order to cope with both occlusions of the lower part of the body and various poses. \blue{Multiple different DPM detectors based on HOG features are used to deal with deformable body pose (as for example sitting, bending etc) in \cite{DPMMultiple}}. Jafari \etal  \cite{aachenTracker} also apply an upper-body detector based on a depth template, as  described in Section \ref{sub:human_detection}, and Liu \etal  \cite{icipColorHeightHist,Liu201516} detect people based on a model of height of the upper-body. Similarly, Zhang \etal  \cite{zhang_real-time_2013} use a poselet-based detector \cite{poselets} to identify body parts, and Bansal \etal  \cite{Bansal2010} perform matches of several local contours, thus allowing the detection of people in arbitrary poses. 
In \cite{Munoz2008b}, detected candidates are checked against a planar model of a standing person using depth information. Thus, sitting people would be rejected by the human detector.

{\bf Miscellaneous --}
Munaro \etal \cite{munaro_tracking_2012,Munaro2014} distinguish people {in close interaction} based on {the separation of their heads} which needs to be at least 30cm. This constraint is generally easily respected, especially in a public environment. 
Han \etal \cite{han_employing_2012} detect people based exclusively on movement and on their height (see Section \ref{sub:human_detection}). Therefore, motionless people would not be detected. In \cite{Munoz2007}, new people are detected by a face detector, and the authors set the detector to only scan the close range area (0.5-2.5m) to speed-up the process. Tracking is still performed on the full space, but would be initialized only after the person enters this detection region.
Constraints on body pose and other miscellaneous constraints are stated in the last column of Table \ref{tab:practical_applications}.

\subsection{Preprocessing the depth map}
\label{sub:preprocessing}

Depth maps tend to suffer from noise and areas of missing values, whatever the sensor, and result in inhomogeneous point clouds. 
A few of the reviewed works correct these deficiencies before exploiting depth information.

\textbf{Depth map denoising and completion --}
Zhang \etal \cite{zhang_real-time_2013} suppress outliers from the 3D point clouds by removing the points that have only a few neighbors.
{In Galanakis \etal \cite{Galanakis2014}, the overlapping views of the structured light sensors create interferences that add noise to the scene. The authors report that in their {setup this} noise is predominantly on the ground plane and negligible on humans, and {use} an estimate of the ground plane to eliminate any points close to the ground in their ROI selection stage.}
The method proposed by Dan \etal in \cite{Dan2012} detects people based exclusively on their 3D shape in the depth map, and so can underperform when faced with missing depth values. To close the holes in their map, they first apply morphological operations to the binarized depth values, obtained by thresholding the heights above the ground plane, and then a nearest neighbor interpolation is used to recover the depth values in the gaps that were filled in the binary map.

\textbf{Voxel grid filtering --}
Works, such as Jafari et al. \cite{aachenTracker} and Almaz\`{a}n and Jones \cite{Almazan2014}, which consider the number of points in ROI clusters have to take into account the perspective effect that makes the density of {points depend} on the distance to the camera. Munaro \etal  \cite{munaro_tracking_2012} address this issue to produce an homogeneous density of points in the volume space by re-sampling their 3D point cloud before ROI detection and clustering. Thus, they ensure that the number of points in a cluster depends only on the size of the object within it, rather than on a combination of its size and distance to the camera. {In \cite{Vo2014}, Vo \etal reduce their initial search space by subsampling their color and depth images.}

\textbf{Fusion of point clouds --}
Jafari \etal  \cite{aachenTracker} obtain richer 3D point clouds by combining those obtained  over a time window of 5 to 10 frames, using their mobile camera motion, estimated by visual odometry. In \cite{munozsalinaspeople2009}, Mu\~{n}oz-Salinas \etal merge the ground plane representation of overlapping views from several sensors by retaining the points in a global coordinate system that have the highest confidence. Galanakis \etal \cite{Galanakis2014} fuse foreground points of overlapping views in a global coordinate system during their ROIs selection stage.
Note that works, such as \cite{almazan_tracking_2013,Almazan2014}, which fuse foreground points of non-overlapping views, do not require specific manipulation of the point clouds and only need to calibrate their cameras' positions and orientations.

\section{Online resources: benchmark datasets and software} \label{sec:resources}
In this section, we provide an overview of publicly available benchmark datasets and {source code}, with a summarised list provided in Tables \ref{tab:data} and \ref{tab:sw}, respectively. 
\subsection{Dataset resources}
The ETH dataset from \cite{ethData} contains stereo sequences obtained by a pair of AVT Marlin F033C cameras mounted on a mobile platform. Images are acquired at $640\times480$ resolution at 14fps. The corresponding disparity maps are obtained by using the stereo algorithm presented in \cite{stereoAlgorithm}, but are not available for download. The dataset is composed {of} 5 sequences recorded in very busy pedestrian zones, and these have been manually annotated every four frames by labeling only pedestrians that are greater than 60 pixels in height. The groundtruth does not contain tracks IDs, hence only the detectors' performances can be obtained. An example of the ETH stereo data is shown in Figure \ref{fig:ETH_dataset}.

\begin{figure} 
\centering
\includegraphics[width=0.88\linewidth]{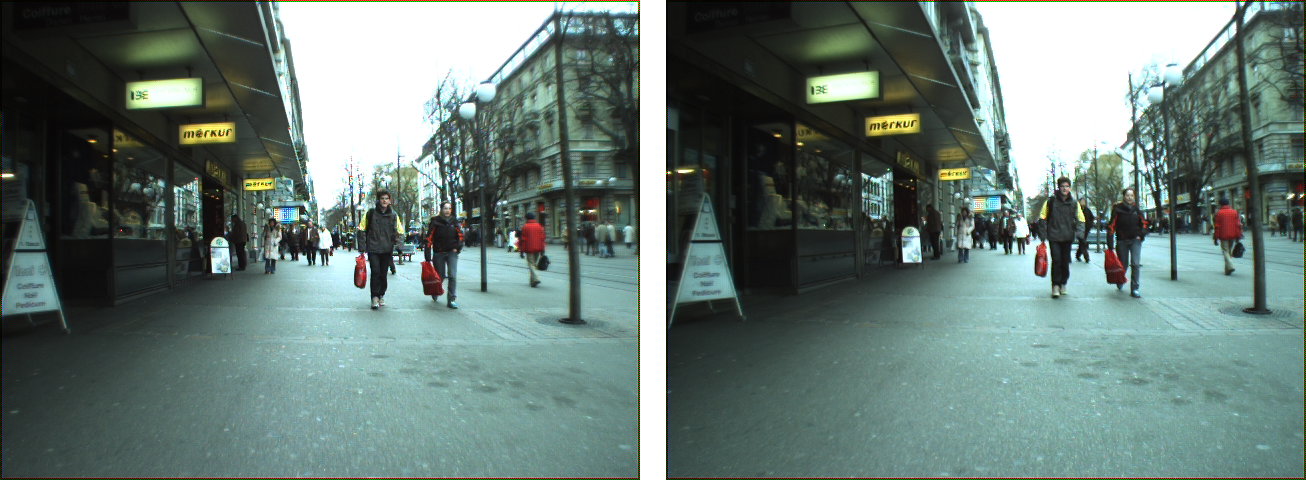}
\caption{{ETH stereo dataset example.}} 
\label{fig:ETH_dataset}
\end{figure}

\begin{figure}  
\centering
\includegraphics[width=0.88\linewidth]{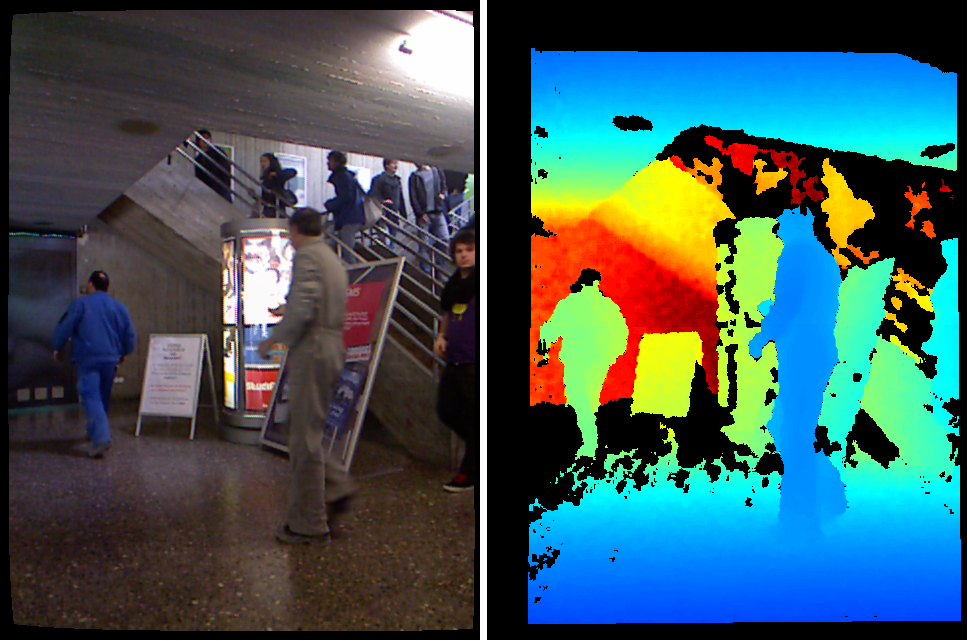}
\caption{{RGB-D UHD {dataset} example.}}
\label{fig:UHD_dataset}
\end{figure}

The dataset presented in \cite{Spinello2011} and \cite{luberIROS11} is obtained by using static cameras, positioned 1.5m high, in a large university hall. We refer to this dataset as {the University Hall Dataset (UHD).} An array of three Kinect devices, with non-overlapping fields of view to avoid IR interferences, is used to record people passing through the university hall.  Due to the Kinect {sensor's} range limitations, depth data is not available beyond a certain range in the hall. The image resolution is $640\times480$, with {synchronised sequences recorded at} 30Hz. This rather small dataset {is composed of three sequences each approximately 1130 frames in length}. There are  3021 instances of people, and 31 tracks have been manually annotated as groundtruth (for detection and tracking). {An example of this UHD data is shown in Figure \ref{fig:UHD_dataset}}. 

The RGB-D tracking dataset presented in \cite{pamiSavarese} {contains two different scenarios captured with Kinects, one static and one mobile. We refer to this dataset as the StanfordRGB-D dataset.}  
The first scenario, the Kinect office, contains 17 sequences with the camera placed 2m high in an office. These videos contain different occlusion scenarios and human poses. The second {scenario, the Kinect mobile, contains 18 sequences with people} performing daily activities in offices, corridors, and hallways. These sequences were recorded with the camera mounted on a mobile platform (a PR2 robot).  In both sets, human positions are hand-annotated (four images every second) with bounding boxes around their upper bodies - hence, both detection data and targets ID are included. Groundtruth odometry information of the camera’s location in 3D space is also available. {An example of the StanfordRGB-D dataset for the static camera scenario is given in Fig. \ref{fig:StanfordRGB-D_dataset}}. 

\begin{figure}
\centering
\includegraphics[width=0.88\linewidth]{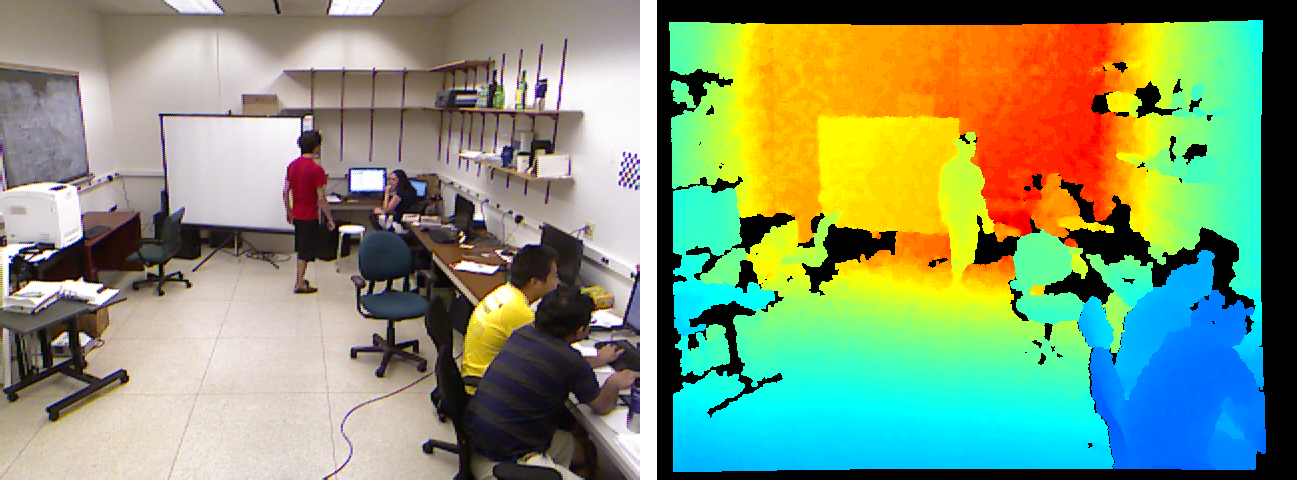}
\caption{{StanfordRGB-D dataset} example.}
\label{fig:StanfordRGB-D_dataset}
\end{figure}

The Kinect Tracking Precision Dataset (KTP) proposed in \cite{munaro_tracking_2012} and \cite{munaro_fast_2014} was acquired with a Microsoft Kinect, at a resolution of $640\times480$ and recorded at  30Hz, on board a mobile platform. It contains 5 different sequences, exhibiting 14766 instances of humans in 8475 frames. Both manually labelled 2D image  and metric groundtruth (for detection and target IDs) are provided, and 3D positions are also available since an infrared marker was placed on every subject's head. Figure \ref{fig:KTP_dataset} shows an example frame from the KTP dataset.

\begin{figure}
\centering
\includegraphics[width=0.88\linewidth]{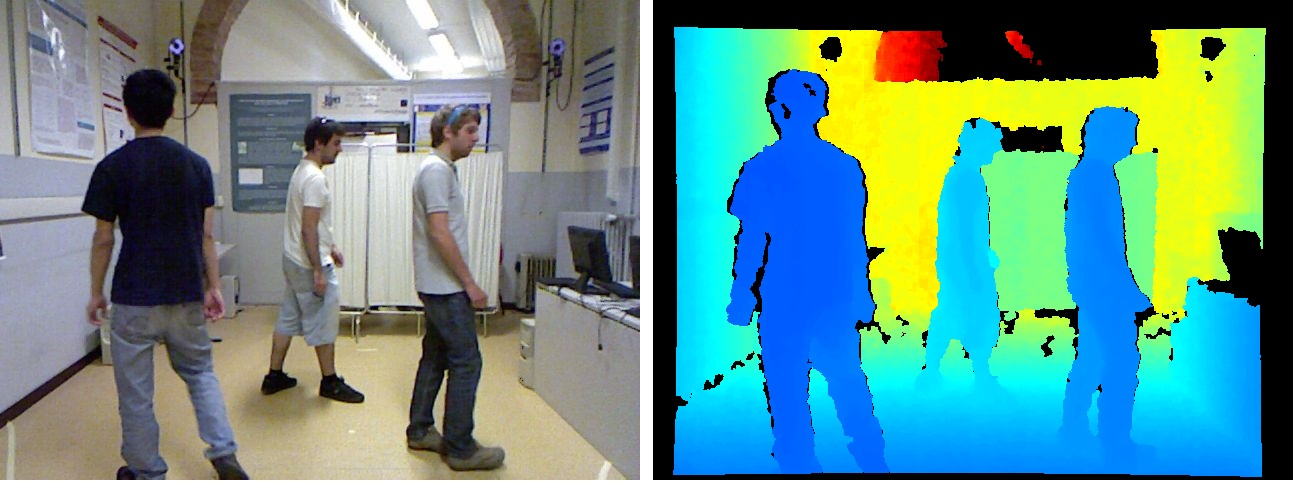}
\caption{RGB-D KTP dataset example.}
\label{fig:KTP_dataset}
\end{figure}

The dataset in \cite{icipColorHeightHist,Liu201516} contains 10 sequences recorded with a Kinect sensor in an indoor (shop) environment, and we refer to it as the SD dataset. The device was mounted at 2.2m high with about a $30^{\circ}$ tilt towards the floor and the sequences were recorded at 30Hz at a resolution of $640\times480$. The groundtruth, produced once every 30 frames, does not contain target ID information, and thus only detection accuracy can be tested. An example of the SD dataset is displayed in Fig. \ref{fig:SD_dataset}.

\begin{figure}
\centering
\includegraphics[width=0.88\linewidth]{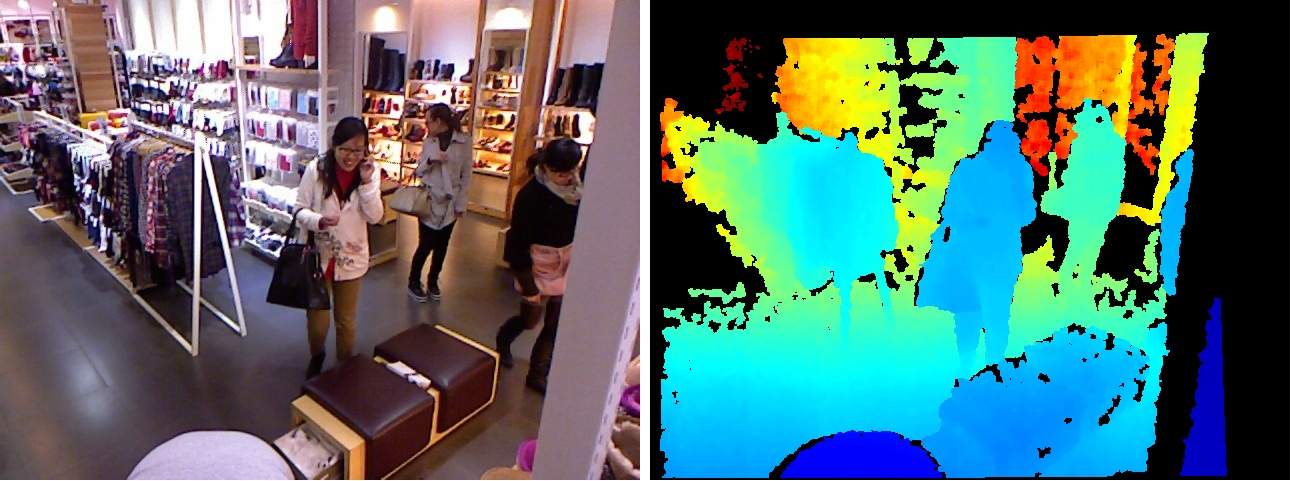}
\caption{RGB-D SD dataset example.}
\label{fig:SD_dataset}
\end{figure}

A recent dataset introduced in \cite{Almazan2014} was obtained using three static Kinect devices, all positioned at about 2m high in a lab with non-overlapping views. We refer to this {multicamera} dataset as the KingstonRGB-D dataset. The sequences contain people moving in the lab, individually or in numbers, with paths crossing {at times}. The dataset comprises six 1000-frame sequences split equally into a training set and a test set. The cameras' calibration matrices and the data to a obtain a wider planar map of the scene are also available. The groundtruth supplies detections and target IDs for all the different views. An example of the KingstonRGB-D dataset is shown in Fig.\ref{fig:Multi_dataset}.

\begin{figure}
\centering
\includegraphics[width=0.88\linewidth]{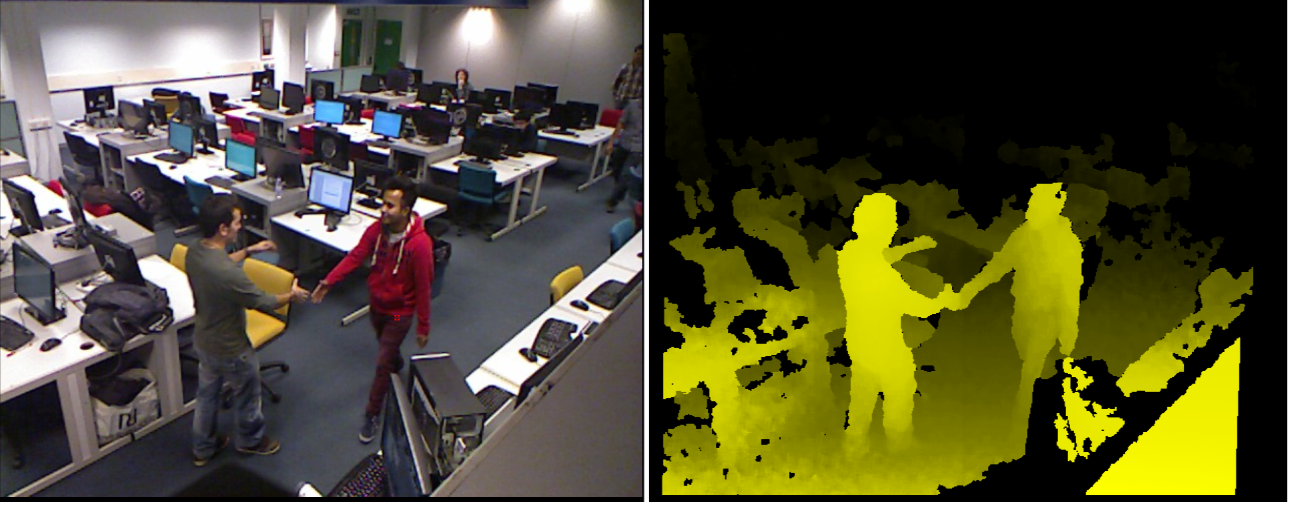}
\caption{RGB-D KingstonRGB-D dataset example.} 
\label{fig:Multi_dataset}
\end{figure}

To recapitulate, only the ETH dataset \cite{ethData} is based on stereo data, while the others presented here 
have all been recorded using the Kinect 
and hence contain only indoor scenes. As also highlighted in Section \ref{subsec:applicationsREQ}, in most of the proposed approaches the position of the camera facilitates the acquisition of frontal views of the moving human. 
Only in the dataset presented in \cite{pamiSavarese}, \cite{icipColorHeightHist,Liu201516} and \cite{Almazan2014} is the camera placed close to the ceiling, giving a top-oblique view of the scene.
This setup yields a more realistic example of a typical surveillance {camera location}.

\begin{table*}
\caption{RGB-D benchmark datasets - In all cases resolution = $640\times480$ and frame rate = $30$Hz (except ETH \cite{ethData} = $14$Hz)}
\label{tab:data}
\centering
\vspace{0.2cm}
\resizebox{0.95\textwidth}{!}{

\begin{tabular}{>{\centering}p{2.5cm}|>{\centering}p{2.5cm}|>{\centering}p{2.9cm}|>{\centering}p{4.2cm}|>{\centering}p{3.9cm}}
& {\bf Device} & {\bf \# Sequences}
{\bf \# Frames} &  {\bf Groundtruth} & {\bf Comments} \tabularnewline
\hline \hline 
ETH
\cite{ethData} & Stereo device (AVT Marlins F033C) & 8

5017 & YES

manually annotated detection only & Minimum pedestrian size 48 pixels, Calibration and odometry data available\tabularnewline
\hline 
UHD
\cite{luberIROS11}, \cite{Spinello2011}  & Multiple Kinect 1 

static & 3

1130 frames per sequence & YES 

Manually annotated detections and 31 people tracks & Part of the scene out of depth range\tabularnewline
\hline 
StanfordRGB-D
\cite{pamiSavarese} & Kinect 1

static and mobile & 17 (static)

18 (moving)

4500 frames per sequence on average & YES

manually annotated four images per second, detections and tracks& Camera positioned 2m high for the static sequences. Groundtruth odometry
of camera location available\tabularnewline
\hline 
KTP
\cite{munaro_fast_2014}, \cite{munaro_tracking_2012} & Kinect 1

static and mobile  & 5

8475 & YES

manually annotated and infrared marker groundtruth, detections and tracks & Device placed at robot level, sequences with different complexity \tabularnewline
\hline 



SD
\cite{icipColorHeightHist,Liu201516} & Kinect 1

static  & 10 & YES

manually annotated one image per second, only detections & Camera positioned 2.2m high, $30^\circ$ inclination. Cluttered and crowded
scenes\tabularnewline

\hline 
KingstonRGB-D
\cite{Almazan2014} & Multiple Kinect 1 

static & 6

1000 frames per sequence &  YES

manually annotated, detections and tracks  & Cameras positioned around 2m high. Cameras'calibration matrices available\tabularnewline \hline

\end{tabular}
}
\end{table*}

\begin{table*}
\caption{Software resources}
\label{tab:sw}
\centering
\vspace{0.2cm}

\resizebox{0.99\textwidth}{!}{

\begin{tabular}{>{\centering}p{2.6cm}|>{\centering}p{2cm}|>{\centering}p{2.9cm}|>{\centering}p{3.5cm}|>{\centering}p{5cm}}
& {\bf Processing Arch.} & {\bf Processing Rate (fps)} & {\bf Dependencies Requirements} & {\bf  \blue{Availability}} \tabularnewline
\hline \hline
Luber \etal 
\cite{luberIROS11}, \cite{Spinello2011}$^{*}$  & CPU & --- & Eigen2 & \blue{Partial: Depth-based detector module and integration with Kinect not available}\tabularnewline
\hline 
Choi  \etal 
\cite{pamiSavarese}$^{*}$ & CPU+GPU & 5-10 & Opencv, boost & \blue{Partial: Depth based detector module not available}\tabularnewline
\hline 
Munaro \etal 
\cite{munaro_fast_2014}, \cite{munaro_tracking_2012} & CPU & 23-28 Detector only, 19-26 Detector + Track & boost, eigen3,flann,Openni,\blue{PCL} & \blue{Partial: only detector module available, manual initialization of ground plane required. Integrated with PCL} \tabularnewline
\hline 
Jafari \etal 
\cite{aachenTracker}$^{*}$ & CPU+GPU & 24 without HOG-GPU 18 with HOG-GPU & FOVIS, Openni x64, CImg, CUDA, KConnectedComponentLabeler, boost,
eigen3, ImageMagick++  & \blue{Partial: missing modules for stereo data processing  to estimate tracked camera position and projections. GPU-CPU processing to enable far distance detections}\tabularnewline
\hline 
{Munaro \etal 
\cite{openPTrack,MunaroSWJournal}} & CPU & 30 & ROS, PCL  & \blue{Full for live camera network, but no plug and play module to test offline data. Multi-camera and Multi-device support for calibration and synchronization.}\tabularnewline \hline
\multicolumn{5}{l}{\blue{$^{*}$ The authors of the paper were contacted (who responded) to clarify details about their software release.}}
\end{tabular}
}
\end{table*}

\subsection{Software resources}
There are only very few  software resources for RGB-D MHT tracking publicly available. A list of these can be found in Table \ref{tab:sw}.  
The source code {for the method presented in \cite{luberIROS11} is available for Linux platforms - however, it does not provide all the functionalities {described} in \cite{luberIROS11}.  For example, the code corresponding to the detection module based on HOD features (described in Section \ref{sub:human_detection}) has not been released, but the code for the tracking core, based on MHT (Section \ref{sub:sequential_det_track}) and the online adaboost detector (Section \ref{sub:sequential_det_track}) are available and are integrated with a laser range scanner.  
Despite the source code being incomplete, this resource is still very useful as the missing HOD module can be developed by the interested researcher starting from one of the available color based HOG versions and then by using the UHD data to train the classifier. The code can also be easily ported onto a Windows environment as its only dependency, Eigen, is available for both linux and Windows. The authors of \cite{luberIROS11} do not provide details of computational performance of their method.} 

The authors of \cite{pamiSavarese} provide the source code for their tracking module, based on a RJ-MCMC particle filter (Section \ref{sub:coupled_tracking}), but some of their proposed detectors (Section \ref{sub:ROI}), in particular their depth-based silhouette (Section \ref{sub:human_detection}), are not made available or integrated into the main processing loop of their software. Their method also runs on Linux, but we have ported it to Windows as the main dependencies needed to run it, OpenCV and Boost, are available on both platforms.

Munaro and co-workers in \cite{munaro_tracking_2012} and \cite{munaro_fast_2014} have integrated \blue{the detector stage of their tracking methodology} into the Point Cloud Library (PCL) \cite{Rusu_ICRA2011_PCL}. This integration with such a widely used library, is one of the main advantages of this source code as it can {be} easily ported to both Windows and Linux. They indicate a processing throughput of 19fps on an Intel i5-520M 2.40GHz CPU and 26fps on an Intel Xeon E31225 3.10GHz CPU; in both cases a 4GB DDR3 memory was used. These remarkable results can be associated  with the specific optimization approaches used, for example, as stated in Section {\ref{sub:preprocessing}}, the algorithm in \cite{munaro_tracking_2012}  dramatically reduces the point cloud size by a subsampling procedure 
and by eliminating ground plane points as described in Section \ref{sub:ROI}. 

{Recently, this software package has been upgraded by Munaro \etal \cite{openPTrack,MunaroSWJournal} to support a multi-camera RGB-D system. The new software library, OpenPTrack, is compatible with Microsoft Kinect and Mesa SwissRanger and can achieve real-time tracking of people at $30$Hz. Each sensor stream independently detects people, while tracking is performed in a central unit by fusing the contribution of all the network nodes. \blue{The detection and tracking software however is not easily accessible as a plug and play module.}
The algorithms presented in \cite{munaro_tracking_2012,munaro_fast_2014,Munaro2014} are included in OpenPTrack.}

Jafari \etal \cite{aachenTracker} provide {the source code for their method which imposes different dependencies as shown in Table \ref{tab:sw}}. The OpenNI library is used as their interface for both the Kinect and Asus Xtion sensors. Their system is based on both a short-range depth based human detector \cite{aachenDetector} running at 24fps on a single CPU and a far-range HOG-based human detector \cite{aachenDetectorHOGGPU} {(see also Section \ref{sub:sequential_det_track})} which must run on a GPU. {Their experimental results were obtained using an Intel i7-3630QM with 12GB RAM and a NVIDIA GeForce GT650m GPU. The main advantage of this software resource is the possibility to activate  the two different detection modules independently. This adaptability offers the opportunity to balance accuracy and processing speed, and the possibility to avoid using modules when not necessary, e.g. the longer range detector in an indoor environment. \blue{Note the system requires calibration and odometry data to operate.}

\section{Comparative evaluations} \label{sec:evaluation}
We now consider how various works have used the datasets introduced in the previous section for {evaluating} their {methods}. 
{Two types of comparative evaluations are presented next - the first attempts to compare different published works on a publicly available dataset (or part of it), while the second presents within-method comparative results by switching one or more of the method's components off.} \blue{Unfortunately, we are not able to compare the results of the software listed in Table \ref{tab:sw} due to the limitations outlined in the previous section.}

\subsection{Inter-Method Comparative Results} Two publicly available datasets have been used by more than one published work, the ETH and the UHD datasets. 

{\bf ETH -} The stereo ETH dataset has been used by  \cite{pamiSavarese,aachenTracker,Bansal2010,munaro_tracking_2012,Ess2009,Bajracharya2009}  to test their specific {methods}, with many utilizing different sequences, and metrics, for evaluation. Bearing this in mind, Table \ref{tab-ethdataset} displays the log-average miss rate (LAMR) \cite{dollarReview} results which is focused on people detection accuracy for the ETH-Bahnhof sequence of the ETH dataset. LAMR is computed by averaging the miss rate vs false positive per image (MR-FPPI) graph in the range $\left[10^{-2}, 1\right]$ in the false positive axis. In particular, we use the reported MR-FPPI results in \cite{pamiSavarese,Bansal2010,munaro_tracking_2012,Ess2009} to extrapolate the LAMR values (second column of Table \ref{tab-ethdataset}). Note, the MR-FPPI results reported in \cite{pamiSavarese,Bansal2010} are not available for the entire range, and for this reason we estimate the Modified\_LAMR (third column of Table \ref{tab-ethdataset}) by considering a smaller interval in the range $[0.056..1]$. The best Modified\_LAMR result is obtained by Choi \etal\cite{pamiSavarese}. 

The sequence `Sunny day' of the ETH dataset is used to test the methods proposed in \cite{aachenTracker,Bansal2010,Bajracharya2009}. The results are reported with graphs of `recall versus false positive per image'. As reported by Jafari \etal \cite{aachenTracker}, 
their method achieves the best results - for example, for a fixed FPPI value of 0.5, their recall rate is $\approx0.85$ which is greater than $\approx0.7$ by \cite{Bajracharya2009} and $\approx0.5$ by \cite{Bansal2010}.

\begin{table}[!h]
\caption{ETH dataset detection results}
\centering
\vspace{0.2cm}

\resizebox{0.46\textwidth}{!}{
\begin{tabular} {c|c|c}
& {\bf LAMR} & {\bf Modified\_LAMR} \tabularnewline
\hline \hline
Ess \etal \cite{Ess2009}  & \textbf{0.645}  & 0.527\tabularnewline
\hline 
Bansal \etal  \cite{Bansal2010}  & -- & 0.612 \tabularnewline
\hline 
Choi  \etal  \cite{pamiSavarese} & -- & \textbf{0.434 }\tabularnewline
\hline 
Munaro \etal \cite{munaro_fast_2014,munaro_tracking_2012} & 0.663 & 0.592\tabularnewline \hline
\end{tabular}
}
\label{tab-ethdataset}
\end{table}

{\bf UHD --} The UDH dataset was used to evaluate the methods proposed in \cite{luberIROS11},\cite{munaro_tracking_2012}\blue{,\cite{DPMMultiple}, and \cite{pamiSavarese} tested with only color-based features}. Tracking performance are reported by considering the CLEARMOT metrics \cite{clearMOT} for which two indexes are given - the {multiple object tracking accuracy (MOTA)} index estimates the tracking error by considering the false negatives, false positives and mismatches, and the {multiple object tracking precision (MOTP)} index which measures how well exact target positions are estimated. False positive, false negative ratios and {identity switches} are also reported. Table \ref{tab:CLEARMOTUHD} shows the results reported in \cite{luberIROS11}, \cite{munaro_tracking_2012,munaro_fast_2014}. While the method proposed by Luber \etal \cite{luberIROS11} guarantees best performance in term of MOTA, FP and FN, the method of Ma \etal allows to dramatically reduce the number of identity switches. \blue{Similar top performance are obtained by Munaro \etal in \cite{munaro_fast_2014} who} state that their poor performance on this dataset is mainly due to mis-detection of people on the staircase sequence as it breaks the flat ground assumption that is central for this approach (as described in Section \ref{sub:ROI}). {When ignoring these mis-detections in the stairs, as well as re-annotating some groundtruth which were believed to be  incorrect,
the authors reported an improved MOTA result of 88.9\%. This result cannot be used for comparative evaluation here as the groundtruth is modified. \blue{For the MOTA metric, the methods presented in \cite{DPMMultiple,pamiSavarese} lead to significantly low scores.}

\begin{table}[!h]
\caption{UHD dataset tracking results}
\centering
\vspace{0.2cm}

\resizebox{\columnwidth}{!}{
\begin{tabular}{c|c|c|c|c|c}
& {\bf MOTP} & {\bf MOTA} & {\bf FP} & {\bf FN} & {ID Sw.} \tabularnewline
\hline \hline
Luber \etal \cite{luberIROS11} & - & \textbf{78.0}\% & \textbf{4.5}\% & \textbf{16.8}\% & 32\tabularnewline
\hline 
Munaro \etal \cite{munaro_tracking_2012,munaro_fast_2014} & \textbf{73.7}\% & 71.8\% & 7.7\% & 20.0\% & \textbf{19}\tabularnewline \hline
\blue{Choi \etal \cite{pamiSavarese} (only color)} & \blue{57.6\%} & \blue{20.2\%} & \blue{20.9\%} & \blue{57.6\%} & \blue{\textbf{1.28}}\tabularnewline \hline
\blue{Ma \etal \cite{DPMMultiple}} & \blue{70.4\% } & \blue{26.9\%} & \blue{13.9\%} & \blue{57\%} & \blue{\textbf{2.1}}\tabularnewline \hline
\end{tabular}
}
\label{tab:CLEARMOTUHD}
\end{table}

{\bf StanfordRGB-D -- }
The StanfordRGB-D dataset has \blue{been} used by its creators in \cite{pamiSavarese}, {and by Liu \etal \cite{Liu201516} and Vo \etal \cite{Vo2014}, to evaluate the detection accuracy of their proposed approaches} to MHT.  {Choi \etal \cite{pamiSavarese}} present their results in terms of  MR-FPPI and the LAMR for the two different scenarios (fixed camera and mobile platform), obtained by averaging across the different sequences. Table \ref{tab:lamrStanford} summarizes the LAMR values, reported in \cite{pamiSavarese}, for the two scenarios: static camera (second column) and moving camera (third column). After the full method in the first row of the table, each row reports the results obtained by turning off one of the detectors (see Section \ref{sub:coupled_tracking}). As shown, the depth cue is the most important for the system, where the LAMR value increases by around 0.25 in both scenarios when this detector is not employed. The HOG based detector is also significant to the final performance of the system, while the impact of the other detectors is {less.} The full system obtains the same LAMR value of around 0.6 for both scenarios. {The recent results obtained by Vo \etal \cite{Vo2014} show that for both scenarios (moving and static cameras) the proposed approach outperforms the results obtained by the RJ-MCMC method in \cite{pamiSavarese}.} {Liu \etal \cite{Liu201516} only report their results in terms of MR-FPPI and thus it is not possible to precisely calculate  Modified\_LAMR values.}  

\begin{table}[!h]
\caption{StanfordRGB-D dataset detection results} 
\centering
\vspace{0.2cm}
\resizebox{0.48\textwidth}{!}{
\begin{tabular}{cc|c|c}
\multirow{2}{*}{} & \multirow{2}{*}{} & \multicolumn{2}{c}{LAMR}\tabularnewline
\cline{3-4} 
Method  &  & Static camera & Mobile camera\tabularnewline
\hline \hline 
\multirow{6}{*}{Choi \etal \cite{pamiSavarese}} \ldelim\{{6}{1mm} & Full & 0.60 & 0.601\tabularnewline
\cline{2-4} 
 & No depth & 0.844 & 0.858\tabularnewline
\cline{2-4} 
 & No Hog & 0.657 & 0.695\tabularnewline
\cline{2-4} 
 & No Face & 0.612 & 0.608\tabularnewline
\cline{2-4} 
 & No Skin & 0.626 & 0.629\tabularnewline
\cline{2-4} 
 & No Motion & 0.592 & 0.637\tabularnewline
\hline 
Vo \etal \cite{Vo2014} &  & \textbf{0.52} & \textbf{0.514}\tabularnewline  \hline
\end{tabular}
}
\label{tab:lamrStanford}
\end{table}

\subsection{Intra-Method Comparative Results} 
Three datasets have been compared on variations of the same method providing comparative results.

{\bf KTP --} 
The KTP dataset was prepared and used by \cite{munaro_fast_2014} to evaluate the tracking performance of their method \cite{munaro_tracking_2012} with the CLEARMOT metrics. In {Table \ref{tab:munaroRes}, we present some of the results reported in \cite{munaro_fast_2014}. The first row shows the results obtained by the algorithm presented in \cite{munaro_tracking_2012} by using all its components, including the sub-sampling strategy described in Section \ref{sub:preprocessing}. This strategy guarantees real-time performance (see Section \ref{sec:resources}) at little loss of performance in comparison to when not subsampling the point cloud (second row). The last three rows contain the results for using different color spaces as input to the color classifier (see Section \ref{sub:sequential_det_track}). The authors claim that the best results are obtained with the HSV color space, especially for the reduction of identity switches.}

\begin{table}[!h]
\caption{KTP dataset tracking results}
\centering
\vspace{0.2cm}
\resizebox{\columnwidth}{!}{
\begin{tabular}{c|c|c|c|c|c}
& {\bf MOTP} & {\bf MOTA} & {\bf FP} & {\bf FN} & {\bf ID Sw.}\tabularnewline
\hline \hline
Full (HSV) & 84.2\% & 85.8\% & 0.7\% & 12.5\% & \textbf{53}\tabularnewline
\hline 
No sub. & 84.2\% & 83.0\% & \textbf{0.6}\% & 15.9\% & 56\tabularnewline
\hline 
Full (RGB) & 84.2\% & 86.1\% & 0.8\% & 12.7\% & 60\tabularnewline
\hline 
Full (CIELab) & 84.2\% & 86.5\% & 0.9\% & \textbf{12.2}\% & 56\tabularnewline
\hline 
Full (CIELab) & \textbf{84.2}\% & \textbf{86.7}\% & 0.9\% & 12.9\% & 65\tabularnewline \hline
\end{tabular}
}
\label{tab:munaroRes}
\end{table}

\blue{As previously mentioned, in \cite{MunaroSWJournal}, Munaro \etal present an extended software library containing the algorithms presented in \cite{munaro_fast_2014} that is able to cope with different depth devices. In  \cite{MunaroSWJournal} they also evaluated the performance of their tracking algorithm by using different devices. They present results for three different sequences 
recorded with both the Kinect (based on structured light technology) and the recent Kinect V2 (based on time-of-flight technology), and  one other time-of-flight device (SR4500). 
The time-of-flight sensors both did better than the first generation Kinect, while Kinect V2 performed better than the SR4500 due to its higher resolution depth representation.}

{\bf SD --} {In \cite{icipColorHeightHist,Liu201516}, the authors first compare color-only to the depth-color detector, reporting the value of the break-even point (i.e. where precision is equal to the recall in the PR curve) of 93\% when the depth-color detector is used, compared to 52\% when the standard color-based HOG detector is employed}. The tracking results presented by the authors in \cite{icipColorHeightHist,Liu201516} are reported in Table \ref{tab:SDresults}. They show how the proposed method based on {depth-color} combination guarantees a {better performance}, for both lost tracks and ID switches, with respect to the proposed tracker relying only on color or depth {solely} to solve the data association problem.

\begin{table}[!h]
\caption{SD dataset tracking results}
\centering
\vspace{0.2cm}
\begin{tabular}{c|c|c}
& {\bf Lost tracks} & {\bf ID switches} \tabularnewline
\hline \hline
Depth-color tracker  & \textbf{13} & \textbf{1}\tabularnewline
\hline 
Depth tracker & 14 & 15\tabularnewline
\hline 
Color tracker & 15 & 6\tabularnewline  \hline
\end{tabular}
\label{tab:SDresults}
\end{table}

{\bf KingstonRGB-D --} This dataset has been used only by its creators in \cite{Almazan2014} to estimate the performance of their tracking method. They evaluate their methods by considering some of the metrics proposed in \cite{otherTrackingMetrics}, i.e. Correct Detected Tracks (CDT), False Alarm Tracks (FAT), Track Detection Failure (TDF), and ID Switches. Additionally, the F1-score metric is used {as a summarizing metric}. The results obtained by the authors in \cite{Almazan2014} are reported in Table \ref{tab:Multiresults} and demonstrate that the proposed tracking strategy based on a color and depth appearance model (described in Section \ref{sub:identification}) is able to outperform {an alternative tracking strategy that uses} depth data only. 

\begin{table}[!h]
\caption{KingstonRGB-D dataset tracking results}
\vspace{0.2cm}
\centering
\resizebox{\columnwidth}{!}{
\begin{tabular}{c|c|c|c|c|c}
 & {\bf CDT} & {\bf FAT} & {\bf TDF} & {\bf F1} & {\bf ID Sw.}\tabularnewline
\hline \hline
Depth/Spatial model & 27 & 18 & 35 & 0.5 & 60\tabularnewline
\hline 
Color+depth & \textbf{40} & \textbf{5} & \textbf{19} & \textbf{0.77} & \textbf{15}\tabularnewline  \hline
\end{tabular}
}
\label{tab:Multiresults}
\end{table}

\section{Challenges}
\label{sec:challenges}

\blue{
In summary, depth data is a fundamental cue that can bring more reliability to MHT methods, but there are many challenges that the vision community needs to address to advance this area further.}

\blue{To start with, it is important that this research community can generate for itself standard and diverse datasets to cover all kinds of application areas (e.g. surveillance, health monitoring, pedestrian tracking, etc) that can help it to evaluate old and new algorithms in a consistent fashion. However, this predicates on researchers to make their data and software more widely available, and report their methodology and processes in a reasonably reproducible fashion.}

\blue{There are still many challenges where depth can be explored further. For example, depth can be a fundamental tool for better (partial) occlusion detection while tracking, so we should expect to see some creative uses of depth information to achieve higher accuracy rates for multiple human tracking - perhaps even in busy scenes depending on the camera viewpoint. Developments on resilient part-based tracking of humans will also help in better occlusion handling.}

\blue{Depth sensors' accuracy is limited to a certain range of distance, hence another important challenge is handling of scale while tracking. The better occlusion and scale handling, the greater the diversity of applications color appearance and depth-based tracking can contribute to. Indoor applications, such as in-home health monitoring may be well served by active sensors, whereas outdoor or longer range applications, such as surveillance monitoring, would be handled by passive sensors. Improvements to the detection range and technology of active sensors will help to overcome shortcomings in scale handling in indoor environments, as already evidenced by the new Kinect V2 compared to the first generation Kinect.}

\blue{
Humans have articulated parts so they will be observed in a variety of poses in various scenarios, compounded by the fact that they also interact with each other. The majority of current works, if not all, track humans while they are walking or standing. A huge challenge lies in tracking people engaged in other activities, e.g. to monitor their routine for health monitoring, or maintaining tracking while they undergo drastic pose changes, e.g. if they bend down, sit down and then stand up, or perform certain prescribed exercises.} 

\blue{Other challenges include more regular issues, such as developing better features and more elaborate adaptive and dynamic methodologies (e.g. such as by applying deep learning techniques).}

\section{Conclusion}
\label{sec:conclusion}
This survey provided an overview of all existing works known to us that fuse RGB and depth data for multiple human tracking. It is a snapshot of the current works in the last few years, along with data and software resources, as well as some comparative results.
MHT is still a relatively young but quickly progressing area where the availability of cheap depth sensors is a huge contributing factor to the regeneration of old, and creation of new, human detection and tracking methods. The analysis and the results reported in the review demonstrates that depth data is fundamental to boost RGB-only MHT methods in terms of both detection accuracy and tracking reliability as depth data introduce very powerful spatial cues (3D shapes and 3D locations) that are also less sensitive to scene illumination conditions. Moreover, the combined color-depth appearance model can be used to describe humans also at region level. 
Further, despite the processing of the additional depth cue, real-time performance can still be maintained, as depth data allows for the significant reduction of the search space for both detector and tracker modules, even when simple heuristic rules are used.

\section*{Acknowledgements}
This work was performed under the SPHERE IRC project funded by the UK Engineering and Physical Sciences Research Council (EPSRC), Grant EP/K031910/1.

\section*{References}
\bibliographystyle{elsarticle-num}

\bibliography{peopleTrackLONG}
%
%

\vfill

\end{document}